\documentclass[11pt]{article}

\usepackage[final]{acl}

\usepackage{times}
\usepackage{latexsym}

\usepackage[T1]{fontenc}
\usepackage[utf8]{inputenc}

\usepackage{microtype}

\usepackage{inconsolata}

\usepackage{graphicx}
\usepackage{pifont}
\newcommand{\cmark}{\ding{51}}   % ✔
\newcommand{\xmark}{\ding{55}}   % ✘
\usepackage{booktabs}   % for \toprule, \midrule, \bottomrule
\usepackage{siunitx}    % for S columns (align decimal points)
\usepackage{caption}    % optional: customize caption
\usepackage{multirow}
\usepackage{listings}
\usepackage{xcolor}

\title{CCD-CBT: Multi-Agent Therapeutic Interaction for CBT Guided by Cognitive Conceptualization Diagram}

\author{
 \textbf{Chang Liu\textsuperscript{1}},
 \textbf{Changsheng Ma\textsuperscript{1}},
 \textbf{Yongfeng Tao\textsuperscript{1}},
 \textbf{Bin Hu\textsuperscript{1}},
\\
 \textbf{Minqiang Yang\textsuperscript{1}},
\\
\\
\\
 \textsuperscript{1}School of Information Science and Engineering, Lanzhou University,
\\
}

\begin{document}
\maketitle
\begin{abstract}
Large language models show potential for scalable mental-health support by simulating Cognitive Behavioral Therapy (CBT) counselors. However, existing methods often rely on static cognitive profiles and omniscient single‑agent simulation, failing to capture the dynamic, information‑asymmetric nature of real therapy. We introduce CCD‑CBT, a multi‑agent framework that shifts CBT simulation along two axes: 1) from a static to a dynamically reconstructed Cognitive Conceptualization Diagram (CCD), updated by a dedicated Control Agent, and 2) from omniscient to information‑asymmetric interaction, where the Therapist Agent must reason from inferred client states. We release \textsc{CCDChat}, a synthetic multi‑turn CBT dataset generated under this framework. Evaluations with clinical scales and expert therapists show that models fine‑tuned on \textsc{CCDChat} outperform strong baselines in both counseling fidelity and positive‑affect enhancement, with ablations confirming the necessity of dynamic CCD guidance and asymmetric agent design. Our work offers a new paradigm for building theory‑grounded, clinically‑plausible conversational agents.
\end{abstract}

\section{Introduction}

\begin{figure}[t]
  \includegraphics[width=\columnwidth]{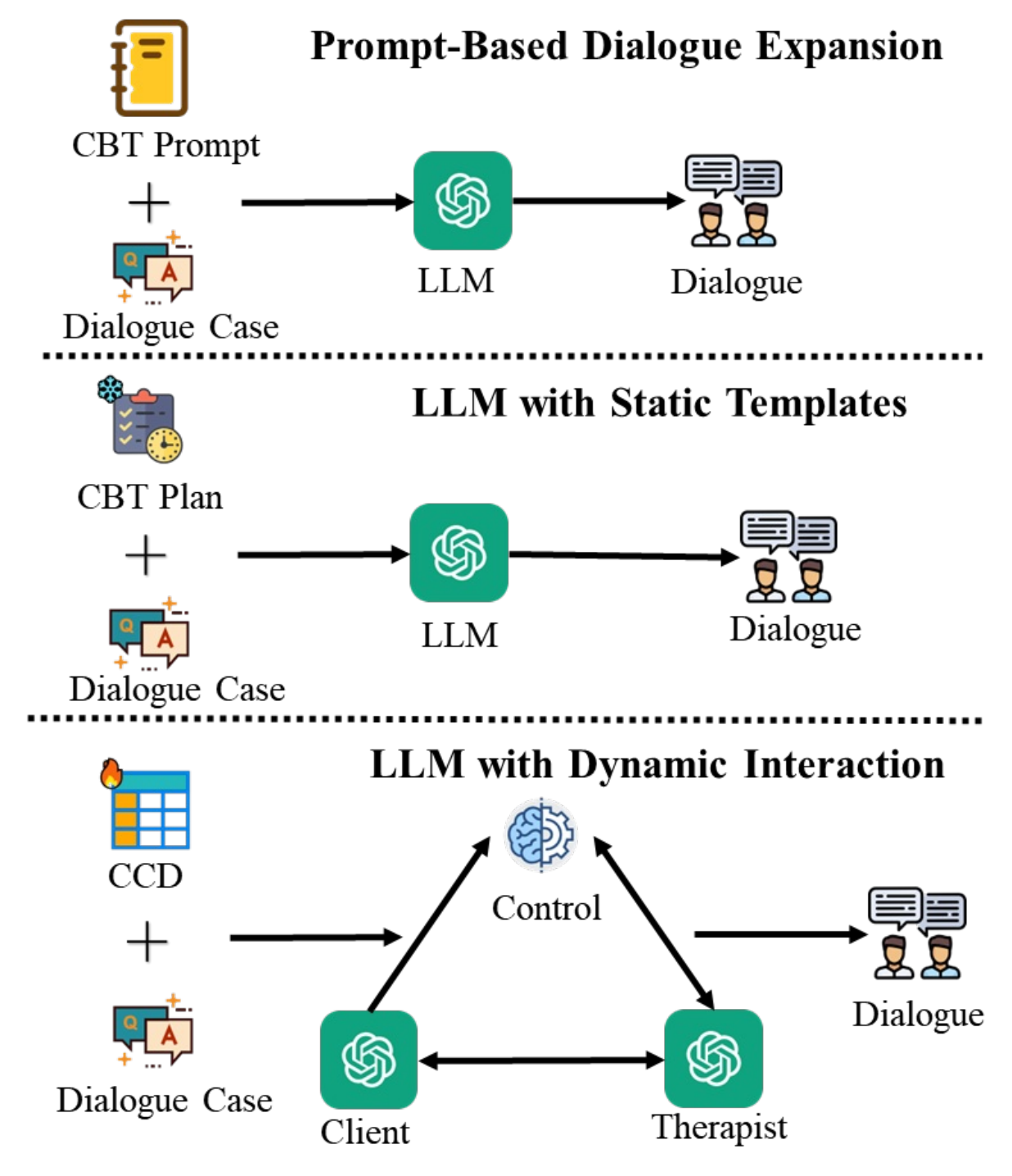}
  \caption{Comparison of CBT simulation paradigms: early prompt-based methods (upper), recent static cognitive modeling (middle), and our dynamic multi-agent framework (lower).}
  \label{fig:intro}
\end{figure}

Mental health disorders, particularly depression, constitute a pressing global burden, exacerbated by a critical shortage of accessible professional therapists~\citep{huang2019prevalence,who2020world}. Large language models (LLMs) offer a promising avenue for scalable mental health support by powering conversational agents~\citep{long2024llms,lawrence2024opportunities}. Among therapeutic approaches, Cognitive Behavioral Therapy (CBT) is a prime candidate for LLM integration due to its structured, evidence-based nature. Initial efforts have integrated psychological theories into LLMs to generate therapist-like responses~\citep{mousavi2021would, das2022conversational, chen2024psychatbot, raile2024usefulness}. However, as illustrated in Figure \ref{fig:intro} (upper), most CBT-oriented systems remain superficial, they rely on prompt-based application of CBT techniques~\citep{na-2024-cbt, xiao-etal-2024-healme} or generate short dialogs in 2- to 3-turns that lack the depth and progression of authentic therapy~\citep{qiu2024smile}.

To achieve greater clinical fidelity, recent work has incorporated structured clinical tools to ground multi‑turn dialogues. Systems like \textsc{CACTUS}~\citep{lee2024cactus} rely on static treatment plans, while DiaCBT~\citep{zhou2025diacbt} explicitly adopts the Cognitive Conceptualization Diagram (CCD)—a core CBT tool for mapping a client’s cognitive‑affective structure~\citep{beck2020cognitive}.
As depicted in Figure \ref{fig:intro} (middle), these approaches represent an advance by using structured cognitive profiles, yet they retain a fundamental limitation: they \textbf{predefine the global cognitive model prior to the session and keep it fixed throughout the interaction}, treating therapy as the execution of a static script.
This contradicts a core tenet of CBT: the therapist \textbf{collaboratively constructs and refines the client’s cognitive model during the session}, starting from limited knowledge and iteratively updating it based on client disclosures~\citep{kuyken2011collaborative,hofmann2012efficacy}.

Furthermore, these systems predominantly \textbf{rely on an omniscient, single-agent simulation paradigm}, where one LLM generates the entire dialogue with full access to both client and therapist internal states~\citep{lee2024cactus, zhou2025diacbt}. As~\citet{zhou-etal-2024-real} critically demonstrate, such omniscient settings create “unrealistically successful” interactions that overestimate social goal achievement and fail to generalize to realistic conditions of information asymmetry. In actual therapy, the therapist has \textbf{no privileged access to the client’s beliefs} and must infer their internal state through active listening and Socratic questioning. The prevailing paradigm thus simulates a distorted reality, lacking the very inferential and adaptive challenges that define therapeutic expertise.

These dual limitations---static cognitive profiles and omniscient simulation---collectively point to a fundamental \textbf{representational mismatch} between existing computational frameworks and the dynamic, dyadic nature of CBT. To bridge this gap, we introduce CCD-CBT, a novel multi-agent framework that instantiates a role-specialized architecture, as visualized in Figure \ref{fig:intro} (lower): a Client Agent (simulating the individual with ground-truth internal states), a Therapist Agent (generating responses based on provided strategy), and a Control Agent (dynamically managing the CCD and therapeutic strategy). This architectural separation enables two critical paradigm shifts: 1. \textbf{From Static to Dynamic CCD:} The Control Agent actively reconstructs the client’s CCD in real-time from the dialogue, transforming it from a static profile into a living state variable that guides the therapy.
2. \textbf{From Omniscient to Asymmetric Simulation:} The Therapist Agent operates without direct access to the client’s ground-truth internal state, relying solely on the Control Agent’s inferred CCD. This enforces the information asymmetry inherent in real therapy, requiring the system to develop genuine clinical reasoning through inference.

Based on this framework, we construct and release \textsc{CCDChat}, a high-fidelity, multi-turn synthetic CBT dialogue dataset. We evaluate models fine-tuned on \textsc{CCDChat} through a dual-pronged strategy incorporating both automatic evaluation using adapted clinical scales and expert human evaluation by licensed CBT practitioners. Results demonstrate that our approach not only outperforms existing baselines in therapeutic skill metrics but also achieves superior outcomes in enhancing clients’ positive affect. Crucially, ablation studies confirm the necessity of both the dynamic CCD guidance and the multi-agent asymmetric setup for generating clinically plausible interactions.
Our contributions are threefold:
\begin{itemize}
    \item We propose CCD-CBT, a novel multi-agent framework that pioneers dynamic CCD reconstruction and information-asymmetric simulation for generating theory-grounded CBT dialogues.
    \item We release \textsc{CCDChat}, a publicly available dataset that captures the collaborative, evolving nature of CBT through our novel simulation paradigm.
    \item We conduct comprehensive evaluations through both automatic and expert human assessments, showing our method generates more effective and authentic therapeutic interactions than current approaches.
\end{itemize}

\begin{figure*}[t]
  \includegraphics[width=0.98\linewidth]{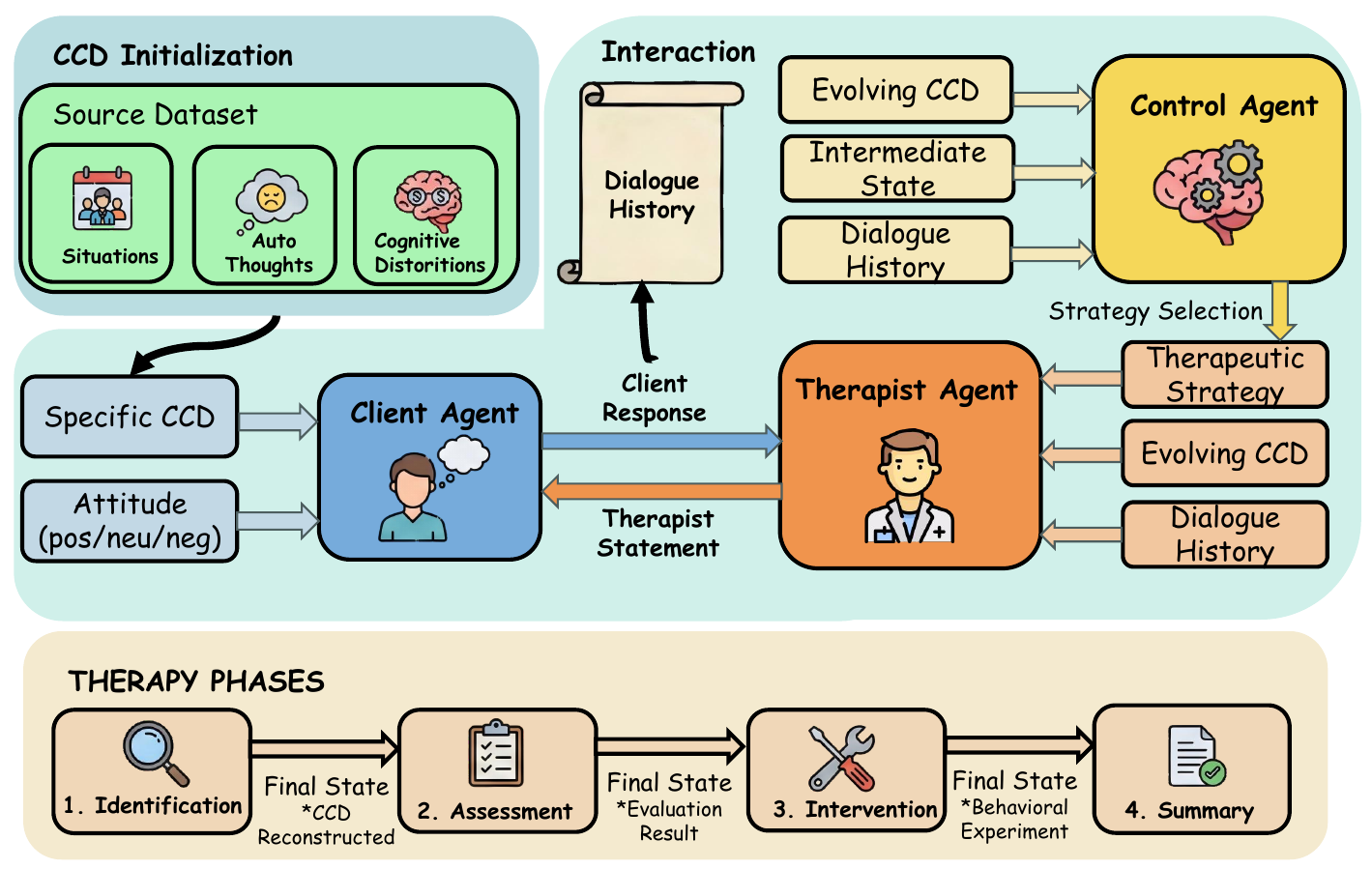}
  \caption {Overview of the CCD-CBT framework. The system consists of a Client Agent, a Therapist Agent and a Control Agent that interact based on the Specific Cognitive Conceptualization Diagram (CCD), guiding the dialogue through four therapeutic phases: Identification, Assessment, Intervention, and Summary. * denotes key states passed to the subsequent phase.}
    \label{fig:framework}
\end{figure*}

\section{Related Work}

\subsection{CBT-based Conversational Systems}

Early LLM-based counseling agents relied on fine-tuning or prompting models on therapy transcripts, producing responses that mimic superficial conversational patterns without grounding in established therapeutic theory or strategy~\citep{liu2023chatcounselor,qiu2024smile}. To incorporate clinical structure, recent work infuses CBT techniques via prompting~\citep{na-2024-cbt} or designs multi‑stage assessment pipelines~\citep{xiao-etal-2024-healme}, yet remains limited to short, non‑interactive exchanges. Truly multi-turn systems have since emerged by generating static treatment plans upfront~\citep{lee2024cactus}, adapting linguistic styles~\citep{xie2025psydt}, or simulating client behaviors by progressively unmasking a predefined cognitive diagram~\citep{kim2025can}. However, these approaches operate from an omniscient perspective, overlooking the clinical reality of information asymmetry~\citep{beck2020cognitive}. Our work addresses this by enabling dynamic, theory‑driven adaptation through a multi‑agent architecture that reconstructs the client's cognitive model in real time.

\subsection{CBT-Related Datasets}

Existing datasets for computational CBT can be categorized into two main types. The first focuses on fine‑grained cognitive‑affective analysis, such as detecting and re‑framing cognitive distortions ~\citep{sharma2023cognitive, maddela-etal-2023-training} or mapping cognitive‑behavioral links \citep{wang2023c2d2}. While these resources support structured reasoning, they do not model multi‑turn therapeutic dialogue.
The second line aims to synthesize counseling‑style conversations. Early datasets like Psych8K~\citep{liu2023chatcounselor} contain transcribed sessions but lack explicit CBT annotation. Recent efforts incorporate CBT techniques more directly: \textsc{CBT‑LLM}~\citep{na-2024-cbt} prompts LLMs to produce single‑turn strategy‑aligned responses,  \textsc{CACTUS}~\citep{lee2024cactus} decouples treatment planning from dialogue generation, enabling diverse scenarios, and PsyDTCorpus~\citep{xie2025psydt} extends interactions by aligning techniques with varied counseling styles. However, these datasets are typically generated via pre‑defined templates or rule‑based simulation, which restricts their ability to reflect the dynamic, co‑constructed progression of real therapy—where each response adapts to the client’s immediate disclosure and the evolving case conceptualization.
Our \textsc{CCDChat} dataset addresses this limitation by using a multi‑agent generation framework that dynamically updates the client’s cognitive model throughout the dialogue, producing interactive counseling sessions.

\section{Method}
We present CCD‑CBT, a multi‑agent framework for generating synthetic CBT counseling dialogues guided by a dynamically updated CCD. Unlike prior script‑based approaches, CCD‑CBT treats the CCD as an evolving internal state that drives clinical reasoning throughout the session. The framework decomposes the therapeutic process into three role‑specific agents: a Client Agent, a Therapist Agent, and a Control Agent. A session begins with a client‑specific CCD and follows the four standard CBT phases: \textit{Identification}, \textit{Assessment}, \textit{Intervention}, and \textit{Summary} ~\citep{beck2020cognitive}. During dialogue, the Control Agent continually refines the CCD and supplies phase‑aware therapeutic strategies to the Therapist Agent, enabling adaptive, multi‑turn interactions that mirror the responsiveness of real therapy. Figure~\ref{fig:framework} illustrates the overall architecture.

\begin{table*}
  \centering
  \small
  \begin{tabular}{lcccccc}
    \toprule
    \textbf{Datasets} & \textbf{Language} & \textbf{Available} & \textbf{Strategy} & \textbf{CBT-based} & \textbf{CCD-guided} & \textbf{Avg. turns} \\
    \midrule
    Psych8k~\citep{liu2023chatcounselor}     & English & \cmark & \xmark & \xmark & \xmark & 1.0   \\
    SmileChat~\citep{qiu2024smile}           & Chinese & \cmark & \xmark & \xmark & \xmark & 10.4  \\
    HealMe~\citep{xiao-etal-2024-healme}            & English & \xmark & \cmark & \cmark & \xmark & 3.0   \\
    CBT-LLM~\citep{na-2024-cbt}                & Chinese & \cmark & \cmark & \cmark & \xmark & 1.0   \\
    \textsc{CACTUS}~\citep{lee2024cactus}             & English & \cmark & \cmark & \cmark & \xmark & 16.6  \\
    PsyDTCorpus~\citep{xie2025psydt}         & Chinese & \cmark & \cmark & \cmark & \xmark & 18.0  \\
    \midrule
    \textbf{\textsc{CCDChat} (Ours)}                   & Chinese & \cmark & \cmark & \cmark & \cmark & 16.2  \\
    \bottomrule
  \end{tabular}
  \caption{\label{datasets-comparison}
    Comparison of existing mental health dialogue datasets.
    \cmark\ indicates support or availability; \xmark\ indicates not supported or unavailable.
  }
\end{table*}

\begin{figure}[t]
  \includegraphics[width=\columnwidth]{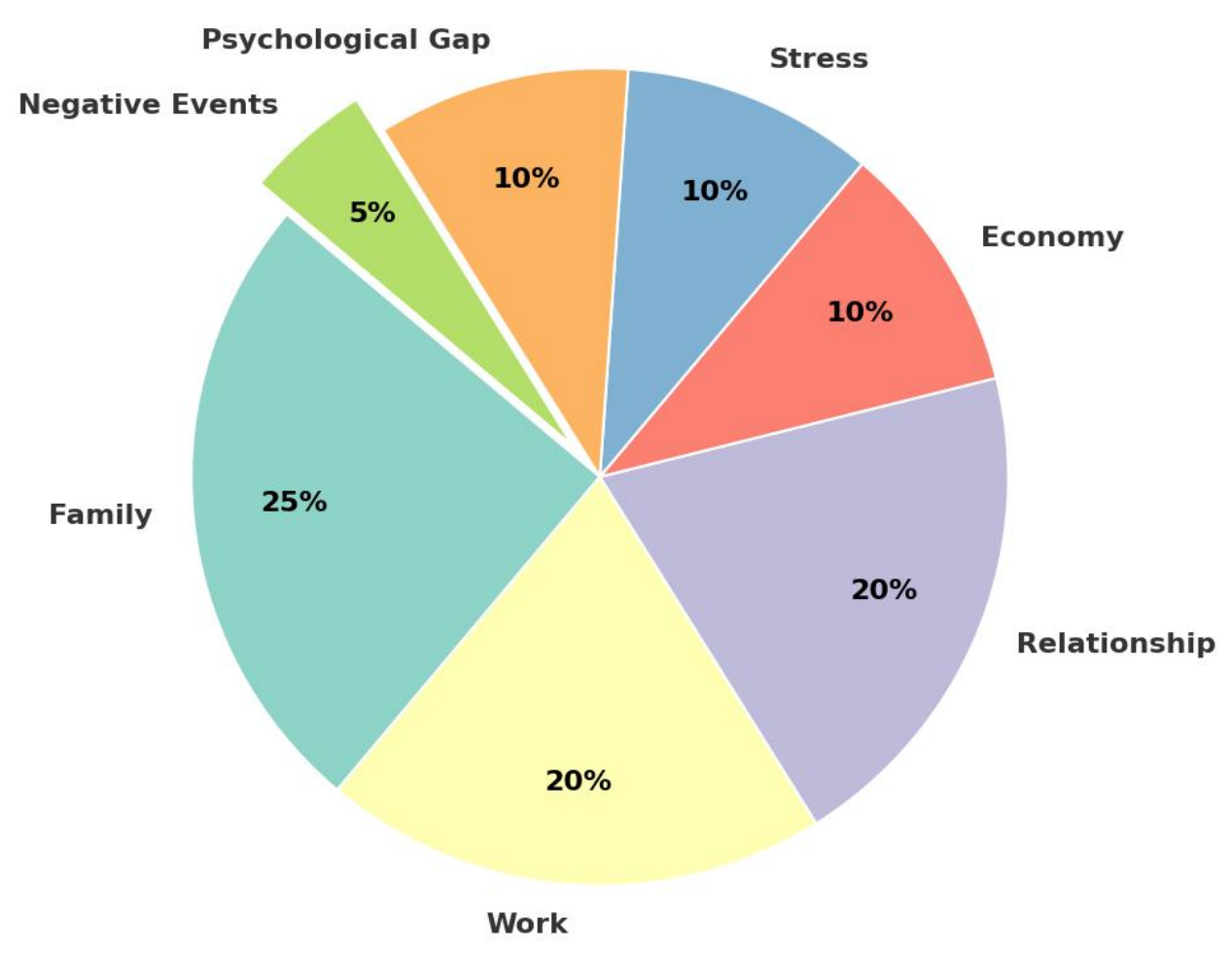}
  \caption{Distribution of situational contexts in CCD-CBT. The dataset covers seven life domains, with Family (25\%) and Work (20\%) being the most frequent triggers for psychological distress.}
  \label{fig:c2d2}
\end{figure}

\subsection{Framework Construction}

\textbf{Source Data \& CCD Construction.} To ground the simulation in clinically plausible cognitive‑affective patterns, we construct 7,500 distinct CCDs using the Chinese CBT resource C2D2~\citep{wang2023c2d2}. C2D2 provides a wide spectrum of everyday scenarios, negative automatic thoughts, and categorized cognitive distortions, as shown in Figure~\ref{fig:c2d2}. Each CCD serves as the ground‑truth cognitive profile of a synthetic client. This structured profile is the seed from which all subsequent interaction emerges; it provides the Client Agent with a consistent internal state and offers the Control Agent a clinical blueprint to reconstruct during the session.\\\\
\textbf{Client Agent.}
The Client Agent embodies the individual seeking therapy. To enhance the realism of client responses, we refer to ~\citep{wang-etal-2024-patient} approach. Its responses are generated autoregressively by prompting an LLM with three inputs: (1) the client’s ground‑truth $\mathrm{CCD}^{\text{client}}_i$, (2) an assigned attitudinal stance $a_i \in \{\text{positive, neutral, negative\}}$ that modulates engagement and resistance, and (3) the dialogue history $\mathcal{H}_{t-1}$. Formally,
\begin{equation}
r_t = \mathcal{LLM}\big( \mathrm{CCD}^{\text{client}}_i,\, a_i,\, \mathcal{H}_{t-1} \big),
\end{equation}
Where $t$ denotes the current dialogue turn.
This design ensures that client utterances remain coherent with their underlying cognitive model while exhibiting realistic, stance‑dependent variation in cooperativeness and emotional tone. Implementation prompts are detailed in Appendix~\ref{app:client_sim}.\\\\
\textbf{Therapist Agent.}
The Therapist Agent generates therapeutic responses by integrating in‑session inferences rather than ground‑truth client information. At each turn $t$, it receives (1) the therapist‑side $\mathrm{CCD}^{\text{therapist}}_i$—a dynamically updated version of the client’s cognitive model reconstructed by the Control Agent, (2) the current CBT strategy $\pi_t$ issued by the Control Agent, and (3) the dialogue history $\mathcal{H}_{t-1}$. The generation is formulated as: 
\begin{equation}
s_t = \mathcal{LLM}\big( \mathrm{CCD}^{\text{therapist}}_i,\, \pi_t,\, \mathcal{H}_{t-1} \big).
\end{equation}
The output $s_t$ is a clinically intentional utterance aimed at techniques such as guided discovery or cognitive restructuring, thereby simulating a therapist who intervenes based on evolving conceptualization, not a fixed script. Prompt details appear in Appendix~\ref{app:therapist_sim}.\\\\
\textbf{Control Agent.}
The Control Agent functions as the clinical reasoning engine of the framework, actively managing the progression of therapy through the four structured CBT phases: \textit{Identification}, \textit{Assessment}, \textit{Intervention}, and \textit{Summary}. It maintains two state variables: the therapist‑side CCD ($\mathrm{CCD}^{\text{therapist}}_i$) and a structured phase tracker $\mathbf{z}_t$. At every turn, it analyses the dialogue history $\mathcal{H}_{t-1}$ and current state to produce the next therapeutic strategy:
\begin{equation}
\pi_t = \mathcal{LLM}\big( \mathrm{CCD}^{\text{therapist}}_i,\, \mathbf{z}_t,\, \mathcal{H}_{t-1} \big).
\end{equation}
Crucially, the agent actively reasons about phase progression. For example, during the \textit{Identification} phase, $\mathbf{z}_t$ follows an 8‑step clinical procedure, as client disclosures arrive, the agent updates $\mathrm{CCD}^{\text{therapist}}_i$ and advances $\mathbf{z}_t$. Upon reaching a phase‑specific terminal state (e.g., completing all eight identification steps), it autonomously triggers a transition to the next phase. This mechanism ensures that the therapy follows a structurally sound yet adaptive trajectory, mirroring the way a human therapist collaboratively builds and refines a case formulation in real time. Full prompt specifications are provided in  Appendix~\ref{app:control_sim}

\subsection{Dataset: \textsc{CCDChat}}

We synthesize the \textsc{CCDChat} dataset by instantiating the complete CCD‑CBT framework: all three agents are implemented with GPT-4o~\citep{hurst2024gpt}. Each dialogue originates from a distinct client CCD and is conducted through the four structured CBT phases. To reflect the depth of real‑world sessions, we generate extended multi‑turn conversations that allow thorough therapeutic exploration. Every CCD is paired with a Client Agent configured under three attitudinal stances, producing an initial set of 3 candidate dialogues.

To ensure high clinical plausibility, we apply a two‑stage quality‑control protocol. First, we discard dialogues exhibiting inconsistent formatting, premature termination, or insufficient turn counts. Second, we conduct an automated competence screening using a proxy of the Cognitive Therapy Rating Scale (CTRS) ~\citep{blackburn2001revised, creed2016implementation}. Six core CTRS items—three for general therapeutic skills, three for CBT‑specific competency—are scored (0–6) per dialogue via LLM‑based judgment. Dialogues whose mean CTRS score falls below a strict threshold of 4 are excluded. After filtering, 98\% of candidates are retained, resulting in the final \textsc{CCDChat} corpus of 4,500 curated, multi‑turn CBT sessions.

As shown in Table~\ref{datasets-comparison}, \textsc{CCDChat} is characterized by its (1) explicit CBT grounding, (2) end‑to‑end CCD guidance, and (3) support for clinically coherent, extended interactions. These properties establish it as a high‑fidelity resource for studying and simulating CBT processes.

\subsection{Model Training}
We fine‑tune a base LLM on \textsc{CCDChat} using Low‑Rank Adaptation (LoRA)~\citep{hu2022lora}. Both the Therapist and Control Agents are optimized in a structured question‑answering format: each turn is framed as a prompted generation task that incorporates role‑specific instructions, the current CCD, dialogue history, and (for the Control Agent) phase state.
For the Therapist Agent. A single LoRA adapter is learned over the entire dialogue trajectory. This encourages the agent to maintain a consistent and empathetic therapeutic style across all counseling phases.
For the Control Agent. We train distinct LoRA adapters for each of the four clinical phases. During inference, the agent’s phase‑specific output acts as a deterministic trigger to either generate the next therapeutic strategy $\pi_t$ or transition to the following phase. This phase‑modular design ensures that strategic reasoning is tightly aligned with clinical objectives while preserving parameter efficiency.
Full hyper‑parameters, training schedules, and prompt templates are detailed in Appendix~\ref{app:training}.

\section{Experiments}
\begin{table*}
  \centering
  \small
  \begin{tabular}{lS[table-format=1.2]S[table-format=1.2]S[table-format=1.2]S[table-format=1.2]S[table-format=1.2]S[table-format=1.2]S[table-format=2.2]}
    \toprule
    \multirow{2}{*}{\textbf{Models}} & \multicolumn{3}{c}{\textbf{General Counseling Skills}} & \multicolumn{3}{c}{\textbf{CBT-specific Skills}} \\
    \cmidrule(r){2-4} \cmidrule(l){5-7}
    & \textbf{Understanding} & \textbf{Interpersonal Eff.} & \textbf{Collaboration} & \textbf{Guided Discovery} & \textbf{Focus} & \textbf{Strategy} & \\
    \midrule
    PsyChat     & 4.50  & 4.36 & 4.48  & 4.20  & 4.19 & 3.77 \\
    SoulChat    & 4.32  & 4.35 & 4.42  & 3.92  & 4.13 & 3.91 \\
    MeChat     & 4.52  & 4.44  & 4.48 & 3.98 & 4.24  & 4.06  \\
    Camel      & 4.78  & 4.52  & 4.68 & 4.50 & 4.39 & 4.16   \\
    SoulChat2.0 & 4.88  & 4.68 & 4.83 & 4.61 & \textbf{4.57} & 4.22 \\
    \midrule
   \textbf{Ours} & \textbf{4.90} & \textbf{4.69} & \textbf{4.87} & \textbf{4.74} & \textbf{4.57} & \textbf{4.36}   \\
    \bottomrule
  \end{tabular}
  \caption{\label{tab:comparison}
    Comparison of general counseling and CBT-specific skills across models.
    Higher scores indicate better performance.
  }
\end{table*}
\begin{table*}
  \centering
  \small
  \begin{tabular}{
    l
    S[input-signs,table-format=+1.2]
    S[input-signs,table-format=-1.2]
    S[input-signs,table-format=+1.2]
    S[input-signs,table-format=-1.2]
    S[input-signs,table-format=+1.2]
    S[input-signs,table-format=-1.2]
  }
    \toprule
    \multirow{2}{*}{\textbf{Models}} &
    \multicolumn{2}{c}{\textbf{Positive Attitude}} &
    \multicolumn{2}{c}{\textbf{Neutral Attitude}} &
    \multicolumn{2}{c}{\textbf{Negative Attitude}} \\
    \cmidrule(lr){2-3} \cmidrule(lr){4-5} \cmidrule(lr){6-7}
    & {\textbf{positive $\uparrow$}} & {\textbf{negative $\downarrow$}} &
    {\textbf{positive $\uparrow$}} & {\textbf{negative $\downarrow$}} &
    {\textbf{positive $\uparrow$}} & {\textbf{negative $\downarrow$}} \\
    \midrule
    PsyChat     & +1.90 & -1.38 & +1.88 & -1.82 & +1.94 & \textbf{-1.84} \\
    SoulChat    & +1.94 & -1.74 & +0.88 & -1.34 & +0.06 & -0.24 \\
    MeChat      & +2.66 & -1.56 & +1.62 & -0.76 & +0.12 & -0.56 \\
    Camel       & +2.22 & -0.46 & +0.40 & -1.90 & +0.18 & -1.52 \\
    SoulChat2.0 & +4.46 & -1.18 & +3.80 & -1.50 & +1.28 & -0.80 \\
    \midrule
    \textbf{Ours} & \bfseries \textbf{+5.98} & \bfseries \textbf{-2.50} & \bfseries \textbf{+5.20} & \bfseries \textbf{-1.92} & \bfseries \textbf{+2.56} & \bfseries -0.96 \\
    \bottomrule
  \end{tabular}
  \caption{\label{tab:panas}
    Results of PANAS across various models.
    This table illustrates the shifts in positive and negative emotions before and after counseling sessions,
    categorized by the client's attitude.
    Higher positive values and lower negative values indicate better performance.
  }
\end{table*}

\subsection{Experimental Setup}
\textbf{Datasets.} We fine‑tune and evaluate models on the \textsc{CCDChat} dataset. For training, we sample 1,500 dialogues for each of the three client attitudinal stances, resulting in a total of 4500 training examples. And we use Qwen2.5-7B-Instruct~\citep{team2024qwen2} as our base model. For evaluation, we randomly select 50 distinct CCDs from the remaining set of unused CCDs, each paired with three different attitudes, resulting in a total of 150 samples as a held-out validation set. The client agent is instantiated using GPT-4o-mini to simulate realistic client responses, enabling assessment of model generalization under clinically plausible conditions.\\\\
\textbf{Baselines.} We compare against two categories of strong baselines. General‑counseling LLMs include: (1) SoulChat \citep{chen-etal-2023-soulchat}, fine‑tuned for empathetic multi‑turn support, (2) PsyChat \citep{10580641}, a client‑centered system that recognizes behavior and selects strategies, (3) MeChat \citep{qiu2024smile}, trained on the real‑world‑derived SmileChat corpus. CBT‑specific models include: (1) CAMEL \citep{lee2024cactus}, a two‑stage agent that first plans then generates CBT‑aligned responses using the \textsc{CACTUS} dataset; (2) SoulChat2.0 \citep{xie2025psydt}, which extends single‑turn dialogues into multi‑turn CBT interactions using PsyDTCorpus.\\\\
\textbf{Evaluation Metrics.} We employ two complementary clinical instruments, both scored automatically by GPT‑4o‑mini to ensure scalability and consistency. (1) CTRS \citep{blackburn2001revised} measures \textbf{clinical competence.} We adapt a streamlined CTRS covering three general counseling skills (Understanding, Interpersonal Effectiveness, Collaboration) and three CBT‑specific skills (Guided Discovery, Focus, Strategy), each rated on a 1–6 Likert scale. Higher scores indicate greater therapeutic competence. (2) Positive and Negative Affect Schedule (PANAS) \citep{watson1988development} evaluates \textbf{emotional outcomes.} It quantifies pre‑ to post‑session changes in client affect via 10 positive and 10 negative emotion items (1–5 scale). We infer the pre‑session state from the initial CCD and predict the post‑session state from the dialogue, computing score differences. Effective counseling should increase positive affect and decrease negative affect.
Detailed prompt templates, scoring protocols, and validation of the automatic rater are provided in Appendix~\ref{app:ctrs} and~\ref{app:panas}.

\begin{table*}
  \centering
  \small
  \begin{tabular}{lS[table-format=1.2]S[table-format=1.2]S[table-format=1.2]S[table-format=1.2]S[table-format=1.2]S[table-format=1.2]S[table-format=2.2]}
    \toprule
    \multirow{2}{*}{\textbf{Models}} & \multicolumn{3}{c}{\textbf{General Counseling Skills}} & \multicolumn{3}{c}{\textbf{CBT-specific Skills}} \\
    \cmidrule(r){2-4} \cmidrule(l){5-7}
    & \textbf{Understanding} & \textbf{Interpersonal Eff.} & \textbf{Collaboration} & \textbf{Guided Discovery} & \textbf{Focus} & \textbf{Strategy} & \\
    \midrule
    w/o CBT           & 4.42 & 4.35 & 4.54 & 4.34 & 4.34 & 4.11 \\
    CBT w/o CCD   & 4.71 & 4.51 & 4.71 & 4.63 & 4.47 & 4.23 \\
    \textbf{CBT w/ CCD}     & \textbf{4.90} & \textbf{4.69} & \textbf{4.87} & \textbf{4.74} & \textbf{4.57} & \textbf{4.36} \\
    \bottomrule
  \end{tabular}
  \caption{\label{tab:ablation}
    Ablation study evaluating the contribution of CBT integration and the control module.
  }
\end{table*}

\subsection{Main Results}
\label{sec:main result}
\textbf{Clinical Competence (CTRS).} We evaluate dialogue models on six CTRS dimensions. As shown in Table~\ref{tab:comparison}, while all models perform reasonably on general counseling skills, scores drop noticeably on CBT‑specific skills, with Strategy being the most challenging dimension.
Our CCD‑CBT model achieves the highest scores in five of the six dimensions, excelling in Understanding (4.90), Interpersonal Effectiveness (4.69), Collaboration (4.87), and Guided Discovery (4.74); it ties for the best score in Focus (4.57). Crucially, it attains a substantially higher Strategy score (4.36), outperforming all baselines and reflecting its stronger capacity for structured intervention and treatment‑plan execution. Among baselines, Camel and SoulChat2.0 show more balanced profiles, yet our model consistently surpasses them, especially in CBT‑specific competencies. These results indicate that our framework delivers superior therapeutic competence, particularly in the core CBT skill of strategic intervention, where existing models struggle most. \\\\
\textbf{Emotional Outcomes (PANAS).} Table~\ref{tab:panas} reports changes in client affect (positive, negative) across three initial‑attitude groups. Our model induces the most favorable emotional shift overall: it yields the largest increase in positive affect for Positive (+5.98), Neutral (+5.20), and Negative (+2.56) clients, and the greatest reduction in negative affect for Positive (-2.50) and Neutral (-1.92) groups. While its reduction for Negative clients is more modest (-0.96), it still exceeds all baselines.
In contrast, baseline models exhibit limited or inconsistent emotional regulation. For example, PsyChat reduces negative affect for Negative clients (-1.84) but shows weak positive‑affect gains; Camel and SoulChat produce smaller changes overall, especially in challenging Neutral/Negative scenarios. Our model not only boosts positive emotions effectively but also achieves competitive mitigation of negative affect, demonstrating a well‑rounded emotional benefit that aligns with therapeutic goals.

Together, these results demonstrate that our framework not only masters procedural CBT skills with superior professionalism, but also drives more clinically meaningful emotional improvement. The gains are especially pronounced in the most challenging dimensions---strategy and negative‑affect reduction---highlighting the value of dynamic CCD guidance and multi‑agent interaction for delivering responsive, evidence‑based digital therapy.

\begin{figure*}[t]
  \includegraphics[width=0.48\linewidth]{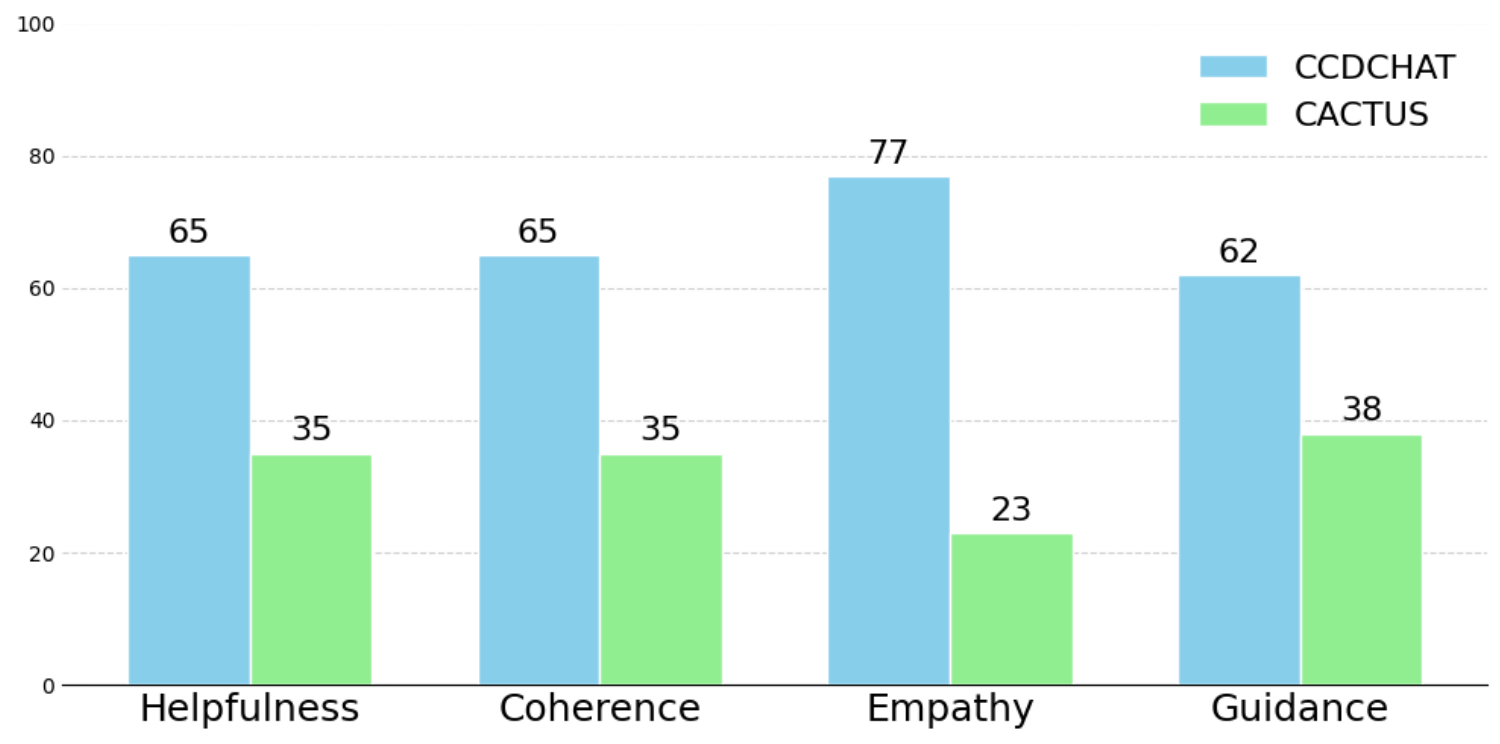} \hfill
  \includegraphics[width=0.48\linewidth]{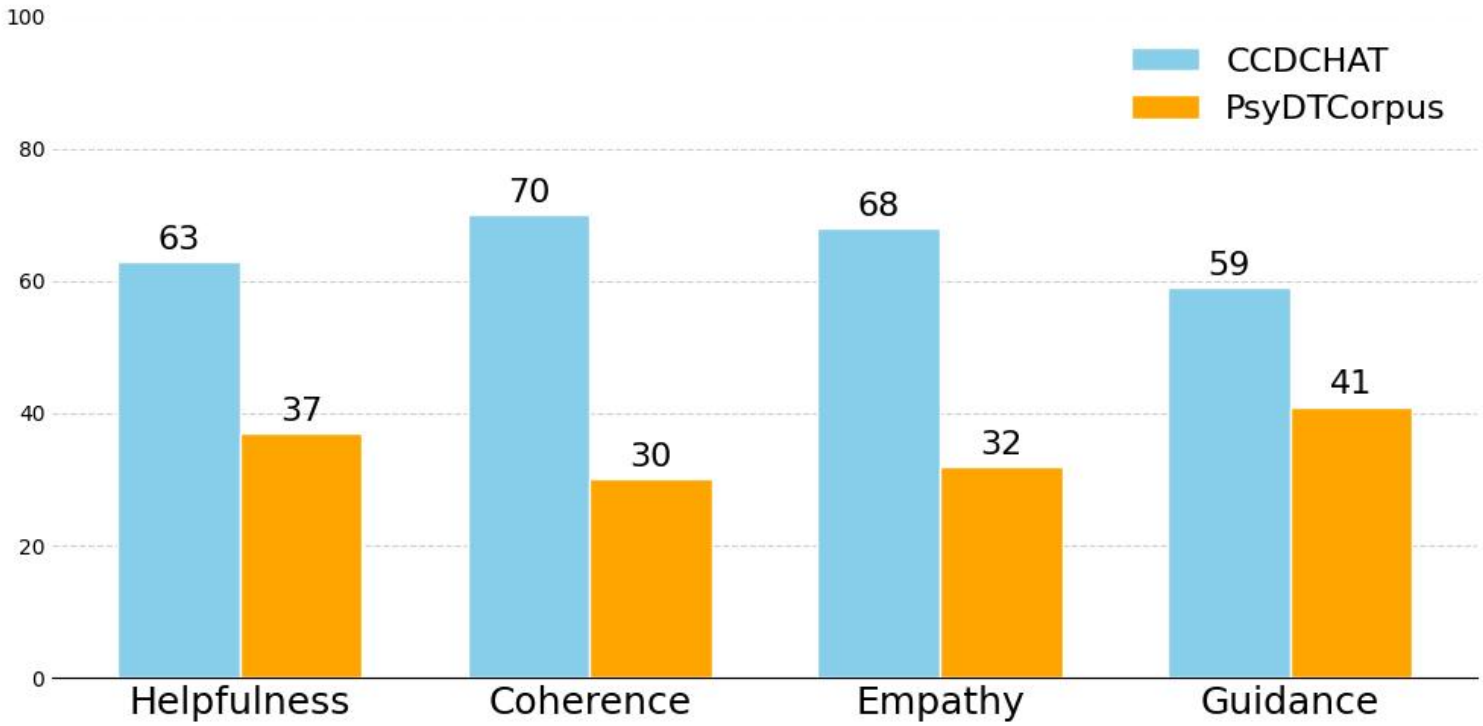}
  \caption {Human evaluation results comparing \textsc{CCDChat} with \textsc{CACTUS} and PsyDTCorpus across four dimensions. In 100 pairwise comparisons per dimension, \textsc{CCDChat} is consistently preferred over both baselines. All differences are statistically significant ($p < 0.05$) according to one-sided binomial tests.}
    \label{fig:humaneval}
\end{figure*}
\subsection{Ablation Studies}
We perform systematic ablations to isolate the contributions of the CCD‑guided planning module and the \textsc{CCDChat} fine‑tuning data.\\\\
\textbf{CCD Ablation.} We compare the full CCD‑CBT model with a variant that retains the same training data but removes the CCD control mechanism (CBT w/o CCD). Results in Table~\ref{tab:ablation} show that ablating CCD guidance leads to consistent performance drops across all six CTRS dimensions. The decline is most pronounced in CBT‑specific skills, especially Strategy (4.36 $\rightarrow$ 4.23) and Guided Discovery (4.74 $\rightarrow$ 4.63), confirming that the dynamic CCD is essential for structured, goal‑oriented therapeutic reasoning. Nevertheless, even without the CCD, the model still outperforms the base LLM in general counseling skills (Understanding 4.71 vs. 4.42), indicating that the \textsc{CCDChat} data alone enhances relational and conceptual alignment. \\\\
\textbf{Data Ablation.} We also evaluate the base LLM without any \textsc{CCDChat} fine‑tuning (w/o CBT). Its scores fall substantially below both the ablated and full models. The gap is largest in CBT‑specific dimensions: w/o CBT trails the full model by 0.25 points in Guided Discovery and 0.23 points in Focus, and achieves the overall lowest score in Strategy (4.11). This confirms that \textsc{CCDChat} provides high‑fidelity behavioral and strategic patterns necessary for clinically competent dialogue.

Overall, the ablation results demonstrate that both components are indispensable and offer complementary benefits: fine‑tuning on \textsc{CCDChat} establishes a foundation of counseling competency, while the CCD module further sharpens CBT‑specific reasoning and strategic coherence. The progressive improvement from w/o CBT to CBT w/o CCD to the full model reflects a clear additive effect of data and structure in building effective, theory‑grounded conversational agents.

\subsection{Dataset Quality Evaluation}
We conduct expert‑based human evaluation to directly assess the clinical authenticity of generated dialogues. Following the protocol of \citet{lee2024cactus}, two licensed CBT practitioners performed pairwise comparisons on 50 randomly sampled dialogues from each of \textsc{CCDChat}, \textsc{CACTUS}, and PsyDTCorpus. Each dialogue was rated across four clinically‑grounded dimensions: Helpfulness, Logical Coherence, Empathy, and Guidance.
As illustrated in Figure~\ref{fig:humaneval}, \textsc{CCDChat} is consistently rated superior to both baseline datasets across all four criteria ($p < 0.05$). While the margin varies across dimensions, the advantage is statistically significant in each case, confirming the superior clinical plausibility of dialogues produced by our CCD‑guided, multi‑agent framework.

This result provides a data‑level explanation for the performance advantages observed in Section ~\ref{sec:main result}: the higher fidelity of \textsc{CCDChat} dialogues furnishes models with more realistic interaction patterns, thereby enhancing both clinical competence and emotional outcomes. Complete evaluation guidelines, inter‑rater agreement statistics, and sample comparisons are included in Appendix~\ref{app:humaneval}.

\subsection{Further Analysis}
\textbf{Client Attitude Effects.} We analyze how client attitude influences simulated counseling quality. Dialogues with positive‑attitude clients receive higher ratings on general counseling skills, likely because cooperative engagement smooths interaction flow. In contrast, neutral‑attitude clients yield the best performance on CBT‑specific skills, suggesting that balanced affect offers a clearer test‑bed for applying structured techniques without emotional interference. Detailed per‑attitude score breakdowns are provided in Appendix~\ref{app:attitude_analysis}.\\\\
\textbf{CCD Reconstruction Fidelity.} A core function of the Control Agent is to dynamically reconstruct the client’s CCD during the \textit{Identification} phase. We evaluate reconstruction accuracy by automatically comparing each inferred CCD component against its ground‑truth source, scoring alignment on a 3‑point scale (1: not accurate $\to$ 3: very accurate). As shown in Table~\ref{tab:ccd_reconstruction}, the reconstructed CCDs achieve high overall alignment (mean = 2.77), with particularly strong matches in Core Beliefs (2.71) and Coping Strategies (2.71). This confirms that the Control Agent can reliably infer clients’ cognitive‑affective patterns from dialogue, a prerequisite for CCD‑guided intervention. The results validate the framework’s capacity for data‑driven conceptualization, underpinning its structured therapeutic reasoning.

\begin{table}
  \centering
  \small
  \begin{tabular}{l
      S[table-format=1.0]
      S[table-format=3.0]
      S[table-format=3.0]
      S[table-format=1.2]}
    \toprule
    \textbf{Component} & 
    \textbf{Not} & 
    \textbf{Slightly} & 
    \textbf{Very} & 
    \textbf{Avg.} \\
    \midrule
    Overall             & 0   & 104 & 346 & 2.77 \\
    Situation           & 1   & 135 & 314 & 2.70 \\  
    Emotions            & 3   & 152 & 295 & 2.65 \\
    Behaviors           & 1   & 108 & 341 & 2.76 \\
    Core Beliefs        & 1   & 129 & 320 & 2.71 \\
    Relevant History    & 2   & 169 & 279 & 2.62 \\
    Coping Strategies   & 1   & 129 & 320 & 2.71 \\
    Intermediate Beliefs& 0   & 167 & 283 & 2.63 \\
    Automatic Thoughts  & 1   & 233 & 216 & 2.48 \\
    \bottomrule
  \end{tabular}
  \caption{\label{tab:ccd_reconstruction}
    CCD reconstruction similarity ratings across components (total $N = 450$).
  }
\end{table}

\section{Conclusion}
We introduced CCD‑CBT, a multi‑agent framework that generates CBT dialogues by dynamically reconstructing a client’s Cognitive Conceptualization Diagram during conversation, moving beyond static and omniscient simulation paradigms. The accompanying \textsc{CCDChat} dataset, validated by clinical experts, enables training of agents that significantly outperform existing baselines in both therapeutic skill and emotional outcomes. Our work demonstrates the value of theory‑guided, adaptive interaction structures for building clinically‑plausible conversational agents for mental health.

\section*{Limitations}
Our study is subject to several limitations. First, our framework simulates therapy as a single extended dialogue, whereas real-world CBT typically spans multiple sessions. Second, the text-only nature of our implementation cannot capture nonverbal cues such as tone and gesture, which are important in real therapeutic interaction. Third, the \textsc{CCDChat} dataset, while clinically grounded, is derived from a limited set of CCDs and is primarily based on Chinese language and cultural contexts, which may affect its generalizability and fairness across diverse populations.

\section*{Ethical Considerations}

\textbf{Data Privacy.} To ensure the highest level of privacy, all dialogues and CCDs within the \textsc{CCDChat} dataset are either synthetically generated, ensuring that no personally identifiable information (PII) is present or can be reverse-engineered from the model’s outputs. \\\\
\textbf{Security.} We acknowledge the inherent stochasticity and potential unpredictability of LLMs. Consequently, the framework is designed to be a supportive instrument used under the rigorous supervision of mental health professionals rather than a standalone counseling entity. Its clinical scope is deliberately restricted to assisting individuals with mild depression and anxiety, ensuring that high-risk or complex cases remain under direct human clinical management. \\\\
\textbf{Bias.} We acknowledge the potential for algorithmic and demographic bias. As our metadata and training corpora are predominantly rooted in a Chinese linguistic and cultural environment, the possibility of population-specific biases remains. The model may inadvertently reflect demographic distributions or social norms prevalent in the source data, which could impact the equity of its responses across different sub-populations.

% \section*{Acknowledgments}

% This document has been adapted
% by Steven Bethard, Ryan Cotterell and Rui Yan
% from the instructions for earlier ACL and NAACL proceedings, including those for
% ACL 2019 by Douwe Kiela and Ivan Vuli\'{c},
% NAACL 2019 by Stephanie Lukin and Alla Roskovskaya,
% ACL 2018 by Shay Cohen, Kevin Gimpel, and Wei Lu,
% NAACL 2018 by Margaret Mitchell and Stephanie Lukin,
% Bib\TeX{} suggestions for (NA)ACL 2017/2018 from Jason Eisner,
% ACL 2017 by Dan Gildea and Min-Yen Kan,
% NAACL 2017 by Margaret Mitchell,
% ACL 2012 by Maggie Li and Michael White,
% ACL 2010 by Jing-Shin Chang and Philipp Koehn,
% ACL 2008 by Johanna D. Moore, Simone Teufel, James Allan, and Sadaoki Furui,
% ACL 2005 by Hwee Tou Ng and Kemal Oflazer,
% ACL 2002 by Eugene Charniak and Dekang Lin,
% and earlier ACL and EACL formats written by several people, including
% John Chen, Henry S. Thompson and Donald Walker.
% Additional elements were taken from the formatting instructions of the \emph{International Joint Conference on Artificial Intelligence} and the \emph{Conference on Computer Vision and Pattern Recognition}.


\bibliography{custom}

\appendix

\section{Cognitive Behavioral Therapy}
\label{cbt}

\subsection{Introduction}
Cognitive behavioral therapy (CBT) is a psychological intervention theory that combines the basic principles of behavioral and cognitive psychology to alleviate symptoms of various mental health conditions, primarily depression and anxiety. CBT focuses on challenging and changing cognitive distortions (such as thoughts, beliefs, and attitudes) and their associated behaviors to improve mood regulation and develop individual coping strategies aimed at addressing current problems.

\subsection{Cognitive Conceptualization Diagram}
\begin{figure}[t]
  \includegraphics[width=\columnwidth]{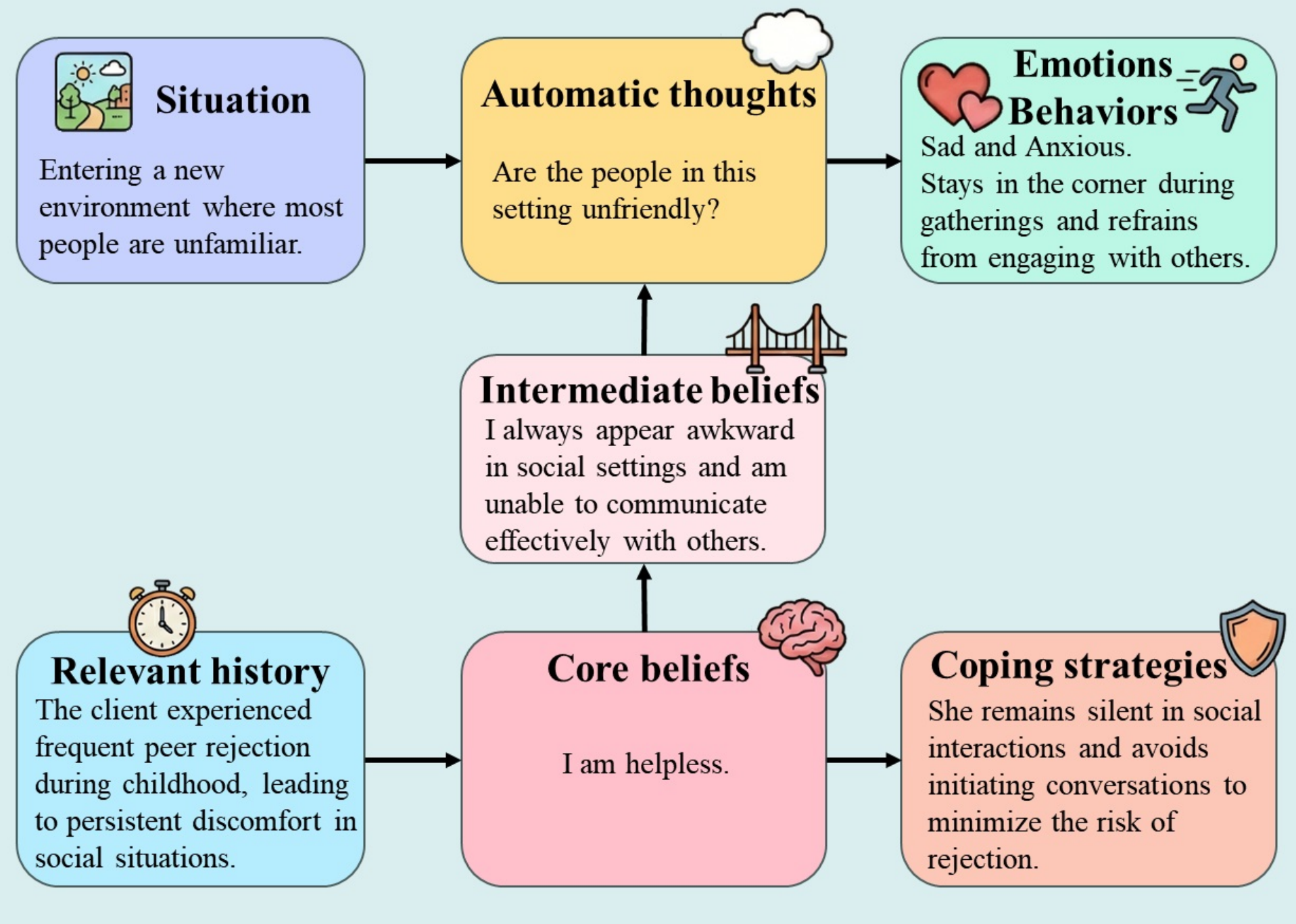}
  \caption{The components of a Cognitive Conceptualization Diagram (CCD) and the relationships between them.}
  \label{fig:ccd_intro}
\end{figure}

The Cognitive Conceptualization Diagram (CCD) is a clinical tool designed to understand and represent an individual’s cognitive patterns and psychological processes. Originally developed by Beck~\citep{beck2020cognitive}, it is widely used in cognitive behavioral therapy (CBT) to guide case formulation and intervention planning. As illustrated in Figure~\ref{fig:ccd_intro}, a complete CCD-based cognitive model comprises eight key components:

\begin{itemize}
\item \textbf{Situation}: The specific event or trigger that initiates the cognitive-emotional sequence.
\item \textbf{Automatic Thoughts}: Immediate, often unconscious thoughts that arise in response to the situation.
\item \textbf{Emotions}: Affective responses (e.g., sadness, anxiety, anger) elicited by automatic thoughts.
\item \textbf{Behaviors}: Bodily sensations or changes (e.g., increased heart rate, tension) associated with emotional states.
\item \textbf{Relevant History}: Significant life events or experiences that have shaped the individual’s mental state and influenced their current beliefs.
\item \textbf{Core Beliefs}: Deeply held, global beliefs about oneself, others, or the world (e.g., “I am unlovable”).
\item \textbf{Intermediate Beliefs}: Conditional assumptions, rules, and attitudes that mediate between core beliefs and automatic thoughts (e.g., “If I please others, I will be accepted”).
\item \textbf{Coping Strategies}: Maladaptive coping behaviors aimed at managing distress linked to core beliefs (optional but often included in extended formulations).
\end{itemize}

In our study, we adopt seven discrete emotional categories: \textit{sad}, \textit{anxious}, \textit{angry}, \textit{disappointed}, \textit{ashamed}, \textit{guilty} and \textit{jealous}. 
Additionally, we focus on three prototypical negative core beliefs that are central to Beck’s cognitive model: \textit{helpless}, \textit{worthless}, and \textit{unlovable}.

\begin{figure}[t]
  \includegraphics[width=\columnwidth]{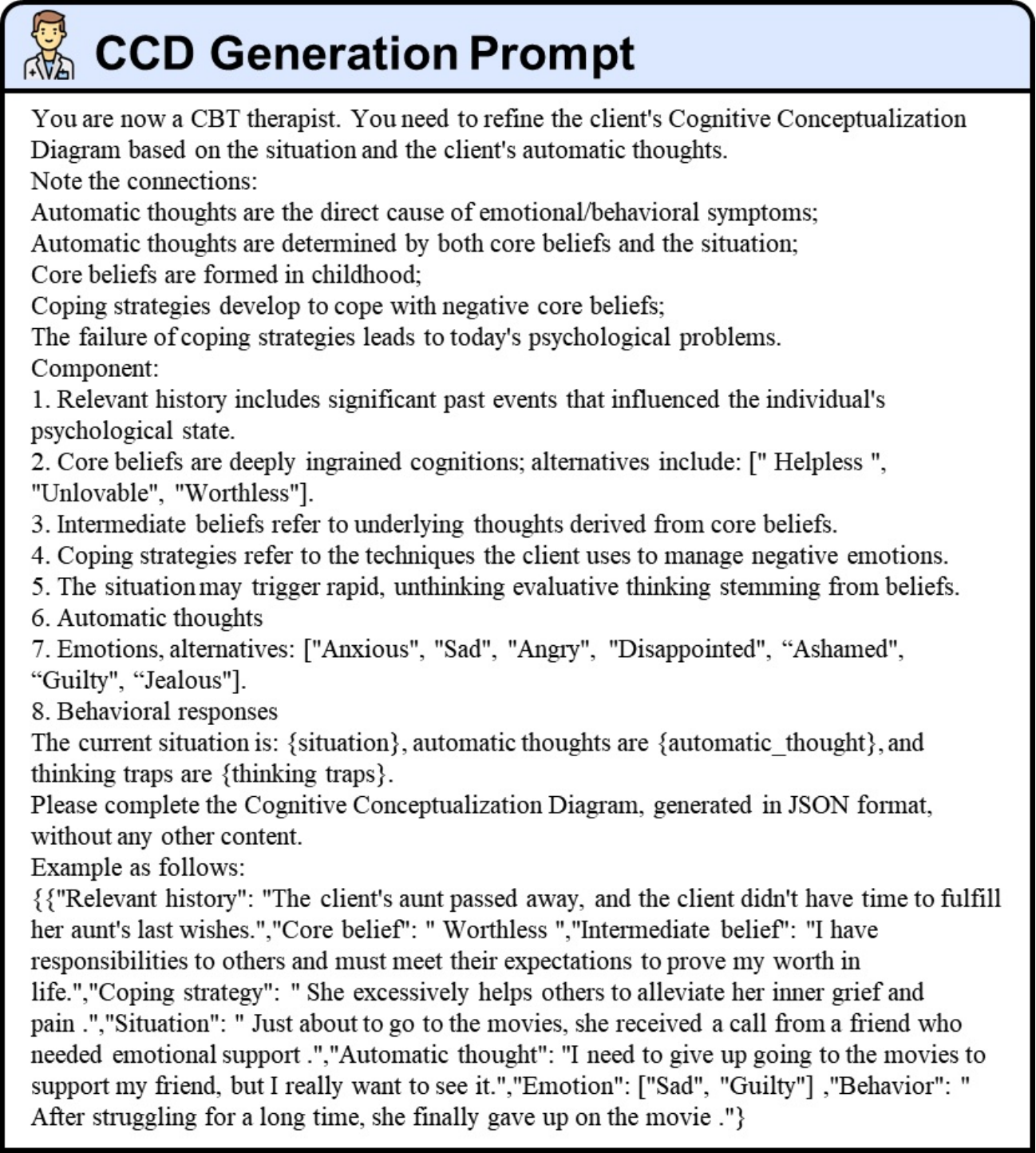}
  \caption{CCD generation prompt.}
  \label{fig:ccd_prompt}
\end{figure}

\section{Details of CCD-CBT}
\label{ccd-cbt}

\subsection{CCD Construction}\label{app:ccdconstruction}
We leverage C2D2~\citep{wang2023c2d2} as the foundational dataset and employ GPT-4o to automatically construct CCDs based on theoretical relationships among CCD components. The detailed prompt used for CCD generation is provided in Figure~\ref{fig:ccd_prompt}.

\subsection{Client Agent}\label{app:client_sim}
\begin{figure}[t]
  \includegraphics[width=\columnwidth]{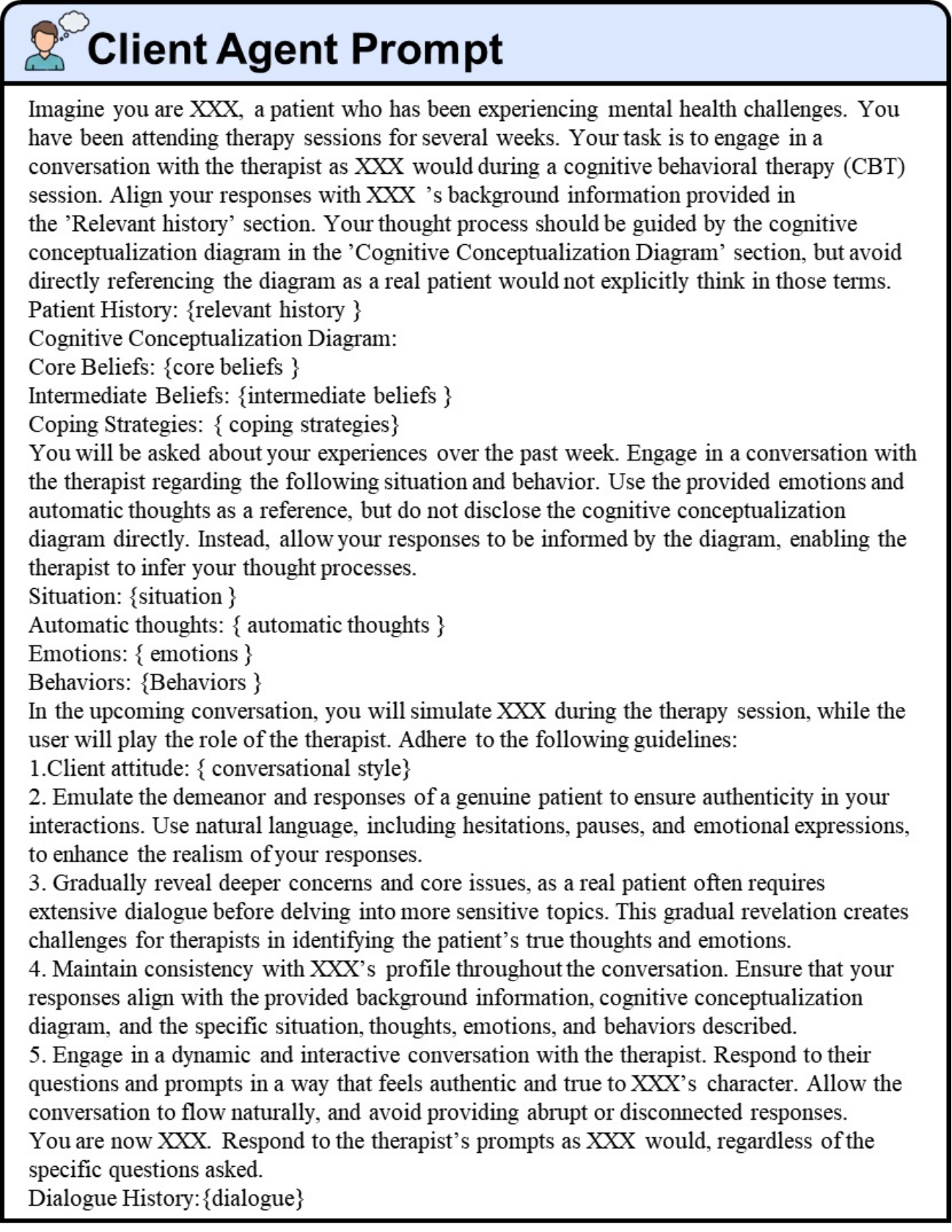}
  \caption{Client Agent prompt.}
  \label{fig:client_prompt}
\end{figure}

We follow the persona design template proposed by~\citet{lee2024cactus} and implement three distinct client attitudes to simulate diverse therapeutic engagement patterns:

\begin{itemize}
\item \textbf{Negative attitude}: The client exhibits resistance or defensiveness, struggles to engage in the therapeutic process, and frequently questions the relevance or efficacy of interventions.
\item \textbf{Neutral attitude}: The client shows occasional willingness to follow guidance and cooperation, but may also display skepticism, hesitation, or mild defensiveness depending on the context.
\item \textbf{Positive attitude}: The client demonstrates high levels of engagement, openness, and cooperation throughout the session, actively participating in therapeutic tasks and discussions.
\end{itemize}
The detailed prompts used to instantiate these attitudes are provided in Figure~\ref{fig:client_prompt}.

\begin{figure}[t]
  \includegraphics[width=\columnwidth]{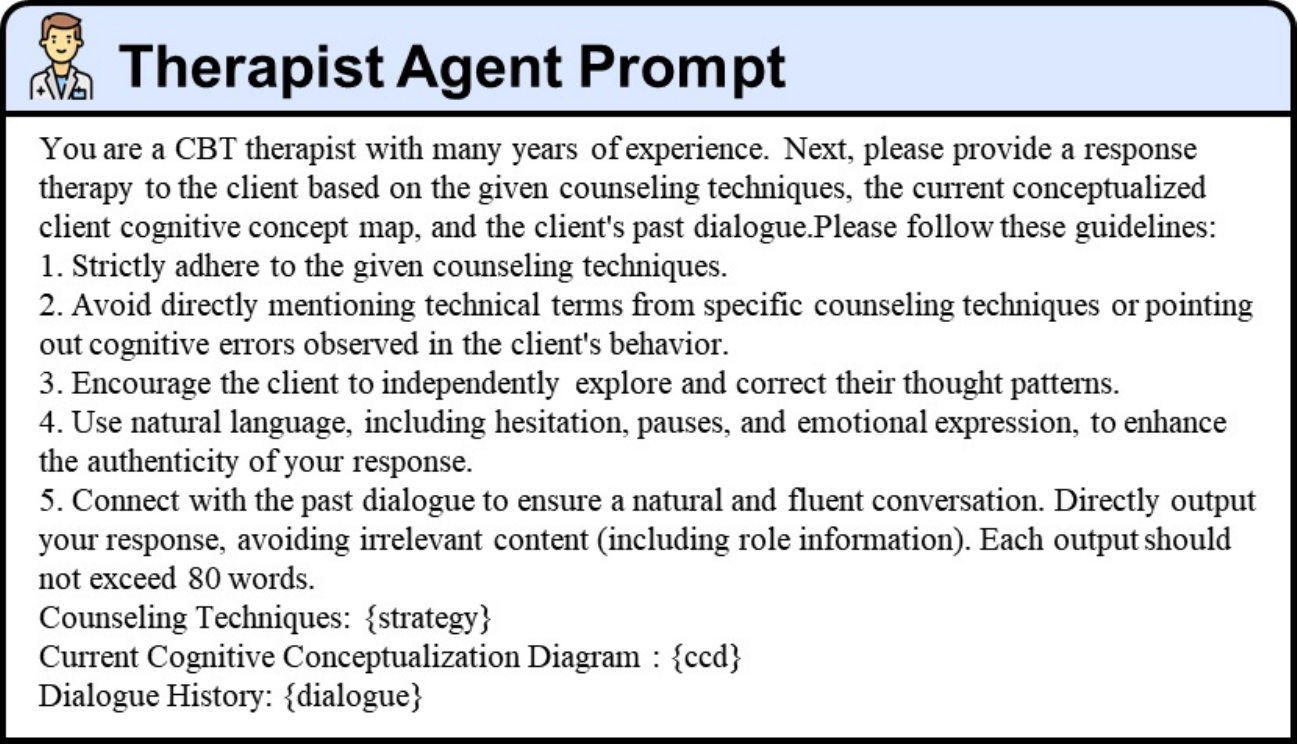}
  \caption{Therapist Agent prompt.}
  \label{fig:therapistprompt}
\end{figure}

\subsection{Therapist Agent}\label{app:therapist_sim}
The therapist agent generates responses based on three key components: (1) the target counseling strategies to be applied, (2) the currently reconstructed CCD, and (3) the dialogue history.The full prompting template is provided in Figure~\ref{fig:therapistprompt}.

\subsection{Control Agent}\label{app:control_sim}
To ensure the stability and logical rigor of the therapeutic process, we decompose the Control Agent into four specialized sub-agents. Furthermore, we implement output validation for both \textit{state} and \textit{strategy}, and impose a maximum turn limit to prevent repetitive loops.\\\\
\textbf{Identification Agent.} 
\begin{figure}[t]
  \includegraphics[width=\columnwidth]{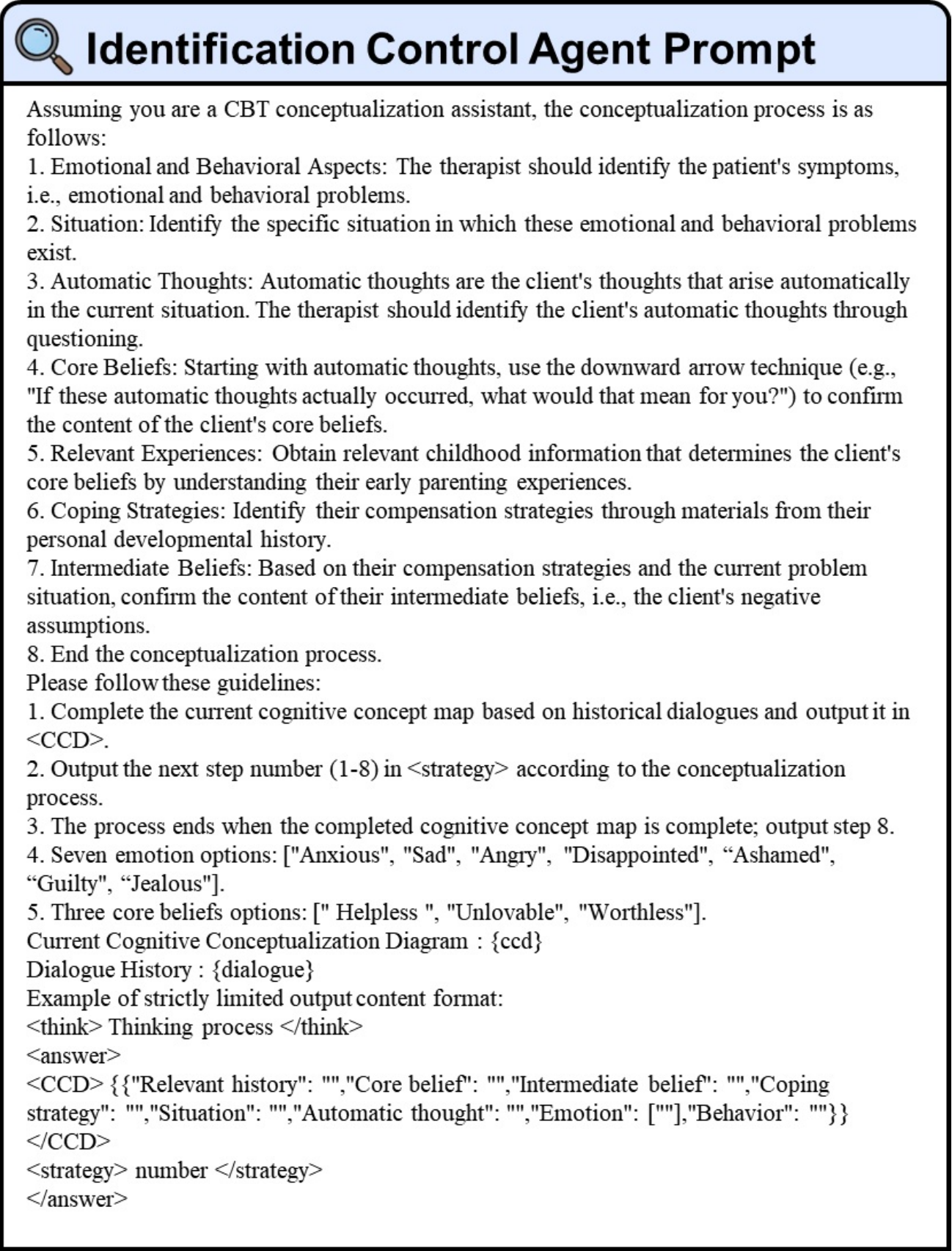}
  \caption{Identification Agent prompt.}
  \label{fig:identificationprompt}
\end{figure}
As illustrated in Figure~\ref{fig:identificationprompt}, this agent functions as a CBT conceptualization assistant. It is responsible for dynamically constructing and updating a structured CCD based on the dialogue history. Following a standardized 8-step cognitive process, it generates the current CCD and determines the next therapeutic phase, ensuring the intervention remains aligned with core CBT principles.
\\\\
\textbf{Assessment Agent.}
\begin{figure}[t]
  \includegraphics[width=\columnwidth]{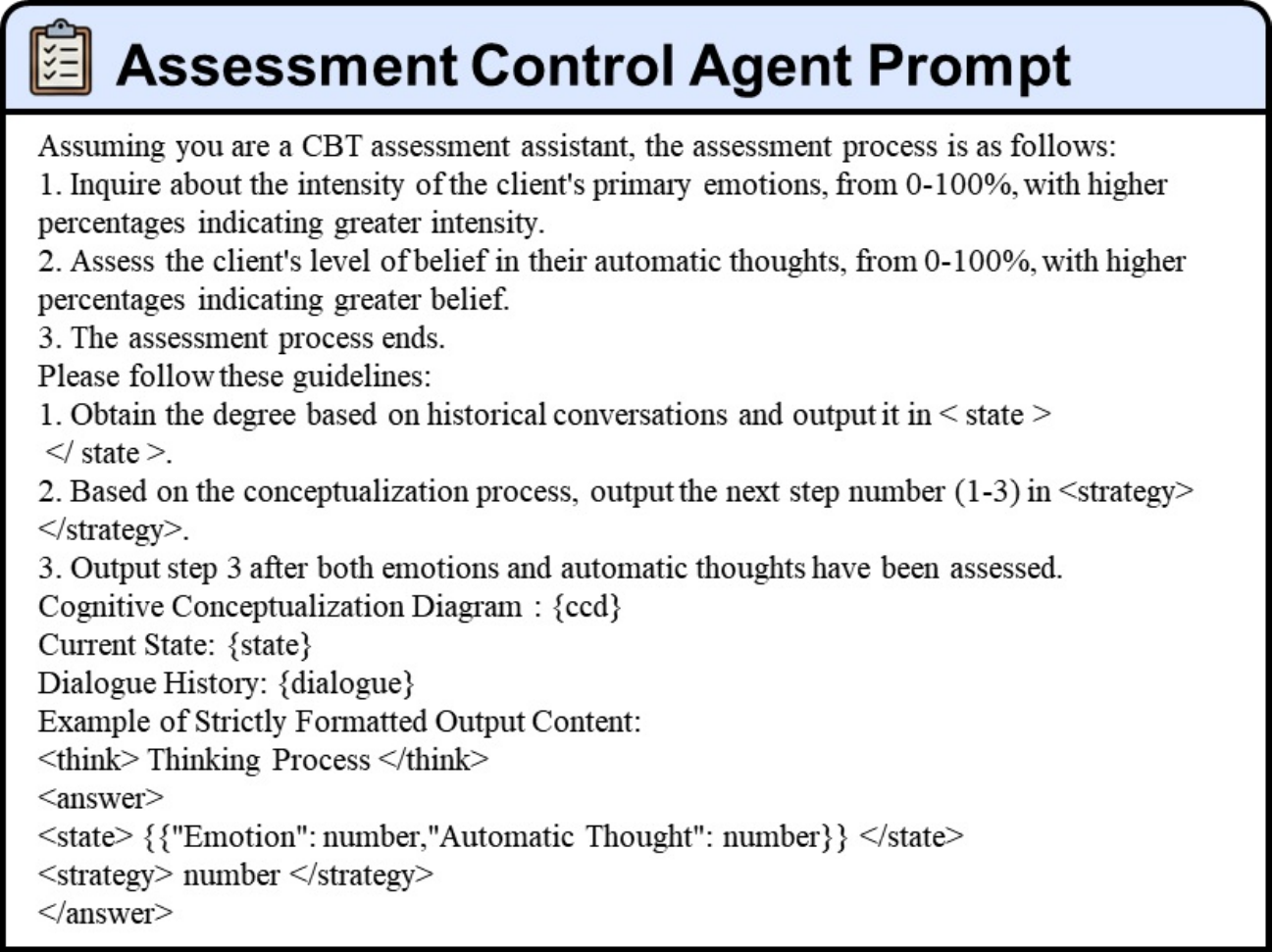}
  \caption{Assessment Agent prompt.}
  \label{fig:assessmentprompt}
\end{figure}
The Assessment Agent functions as a CBT evaluation assistant, employing a Scale Assessment Method ~\citep{beck2020cognitive}to quantify the client's emotional intensity and belief strength in their automatic thoughts. As shown in Figure~\ref{fig:assessmentprompt}, it analyzes the dialogue history to identify linguistic and contextual markers indicative of the primary emotionand the degree of belief in automatic thoughts. The agent outputs the current cognitive-affective state in a structured format, which directly guides the subsequent therapeutic intervention in alignment with CBT principles.
\\\\
\textbf{Intervention Agent.}
\begin{figure}[t]
  \includegraphics[width=\columnwidth]{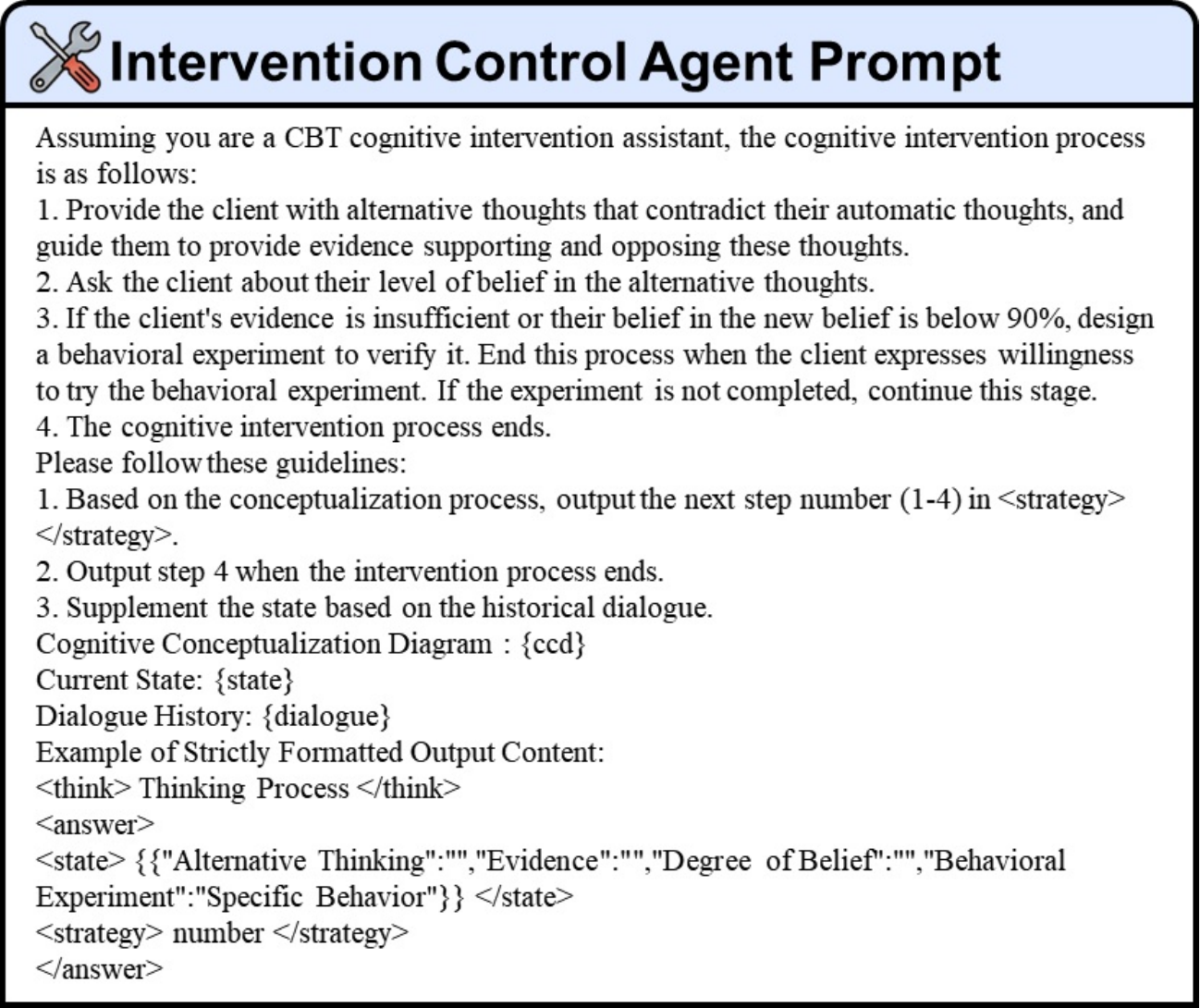}
  \caption{Intervention Agent prompt.}
  \label{fig:interventionprompt}
\end{figure}
The Intervention Agent serves as a CBT cognitive intervention assistant, responsible for guiding clients through structured thought restructuring and behavioral change. As illustrated in Figure~\ref{fig:interventionprompt}, it follows a four-step process: (1) generating alternative thoughts that challenge automatic negative thoughts; (2) assessing the client's level of belief in these alternatives; (3) designing a behavioral experiment when belief remains below 90\%; and (4) concluding the intervention with successful engagement.
Notably, if a behavioral experiment is designed, it is assigned as homework for the client to carry out between sessions. This ensures continuity of therapeutic work and provides empirical data for future assessment and conceptualization refinement.
\\\\
\textbf{Summary Agent.}
\begin{figure}[t]
  \includegraphics[width=\columnwidth]{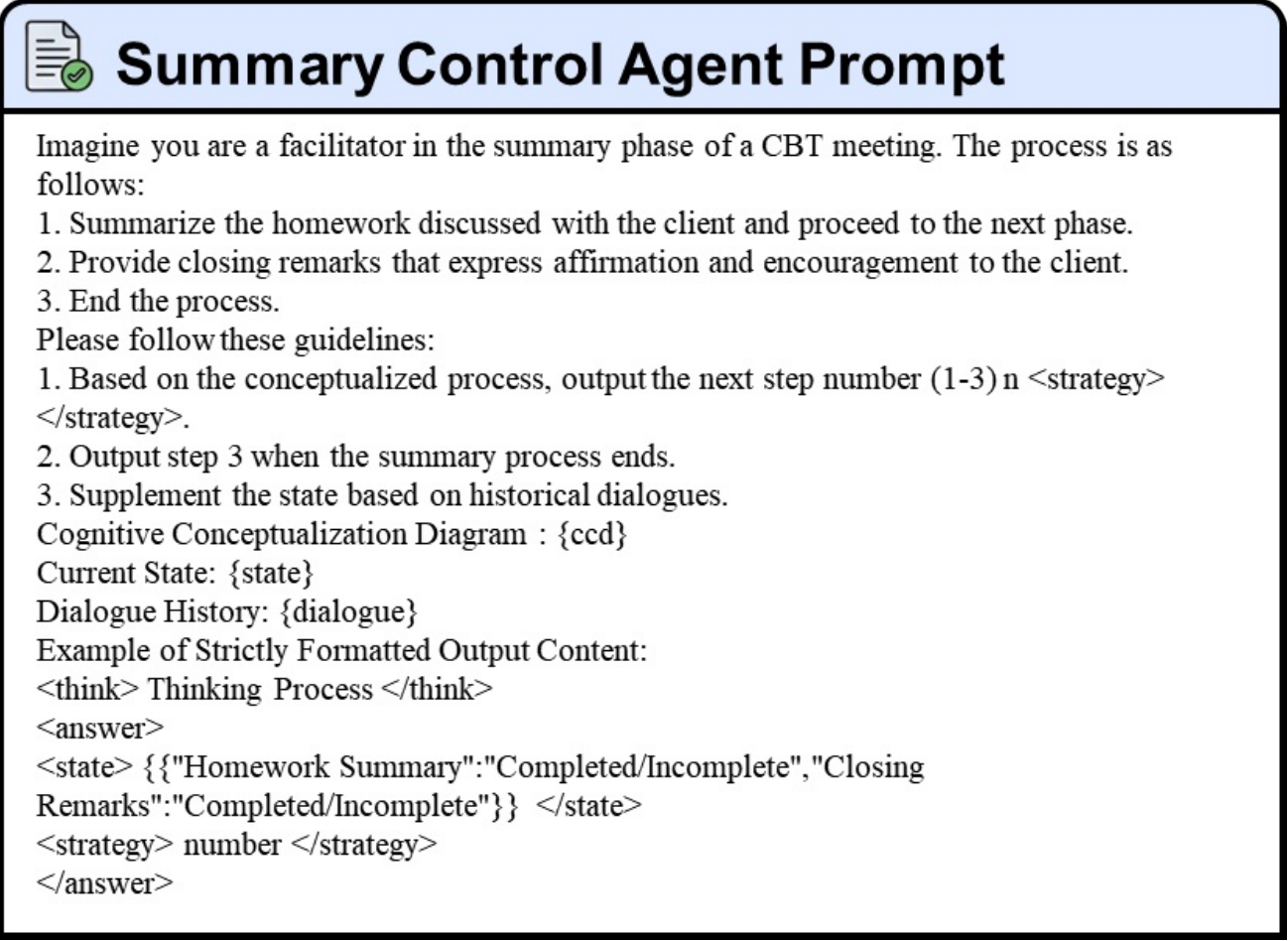}
  \caption{Summary Agent prompt.}
  \label{fig:summaryprompt}
\end{figure}
The Summary Agent functions as a facilitator in the final phase of a CBT session, responsible for consolidating therapeutic progress and reinforcing client engagement. As shown in Figure~\ref{fig:summaryprompt}, it performs three key tasks: (1) summarizing the client’s completed or pending homework, particularly behavioral experiments assigned during the intervention stage; (2) delivering closing remarks that provide affirmation and encouragement; and (3) concluding the session. By explicitly acknowledging the behavioral experiment as a core component of homework, it ensures continuity between sessions and supports the client’s self-efficacy in applying CBT principles outside therapy.

\section{Evaluation Method}
Given that ~\textsc{CACTUS} is in English, we translate it into Chinese using GPT-4o and evaluate both language versions, selecting the higher score for each metric to mitigate potential bias from translation quality.

\subsection{Human Evaluation}
\label{app:humaneval}

\begin{figure}[t]
  \includegraphics[width=\columnwidth]{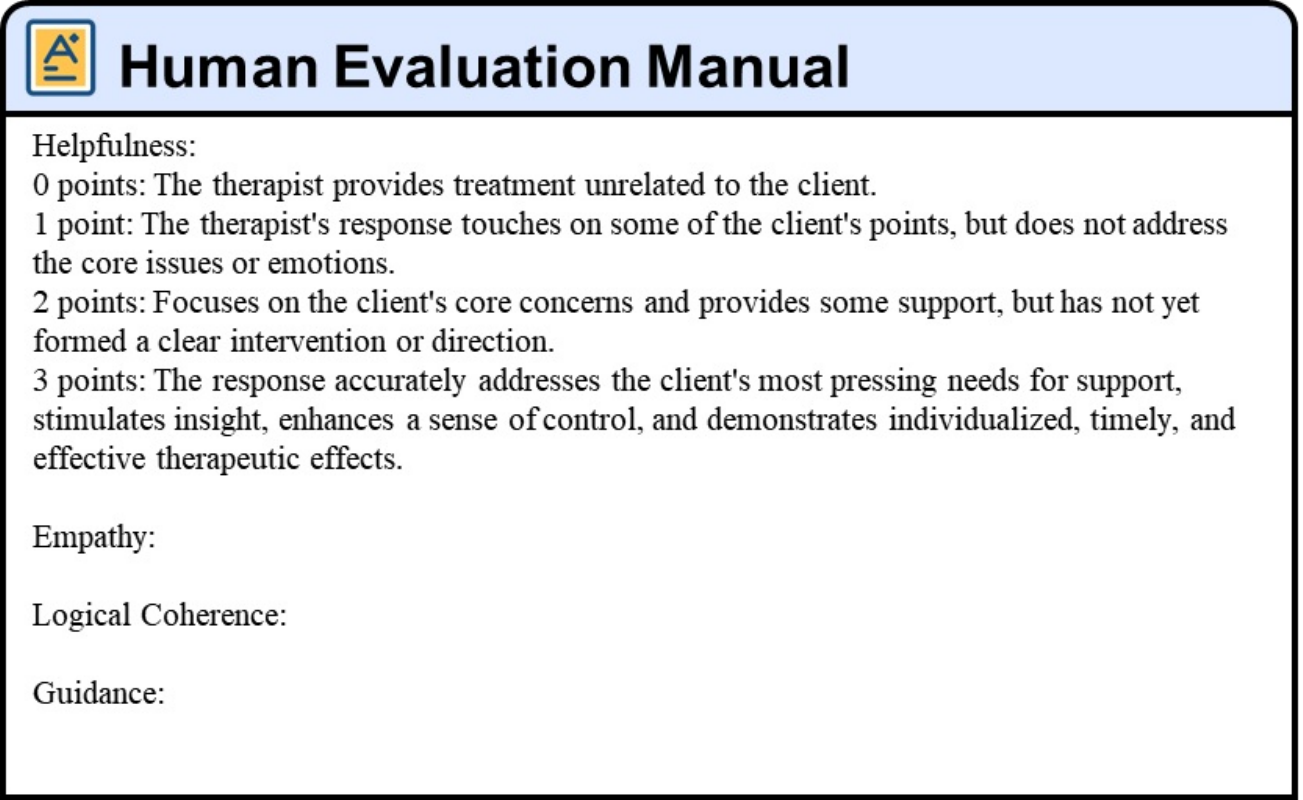}
  \caption{Human Evaluation Prompt.}
  \label{fig:humanmanual}
\end{figure}

To ensure clinical validity and therapeutic quality, two licensed CBT practitioners evaluated dialogues using a structured 4-point rubric. The assessment covered four dimensions: \textit{Helpfulness}, \textit{Empathy}, \textit{Logical Coherence}, and \textit{Guidance}. The detailed scoring criteria for \textit{Empathy}, \textit{Logical Coherence}, and \textit{Guidance} follow the definitions in~\citet{lee2024cactus}; the \textit{Helpfulness} criterion was adapted to emphasize the  effectiveness of therapeutic progress within the CCD-CBT framework. Annotators were recruited via a professional crowdsourcing platform and were compensated at a rate of 8 RMB per data item. The full evaluation manual is shown in Figure~\ref{fig:humanmanual}.

\subsection{Cognitive Therapy Rating Scale (CTRS)}\label{app:ctrs}
\begin{figure}[t]
  \includegraphics[width=\columnwidth]{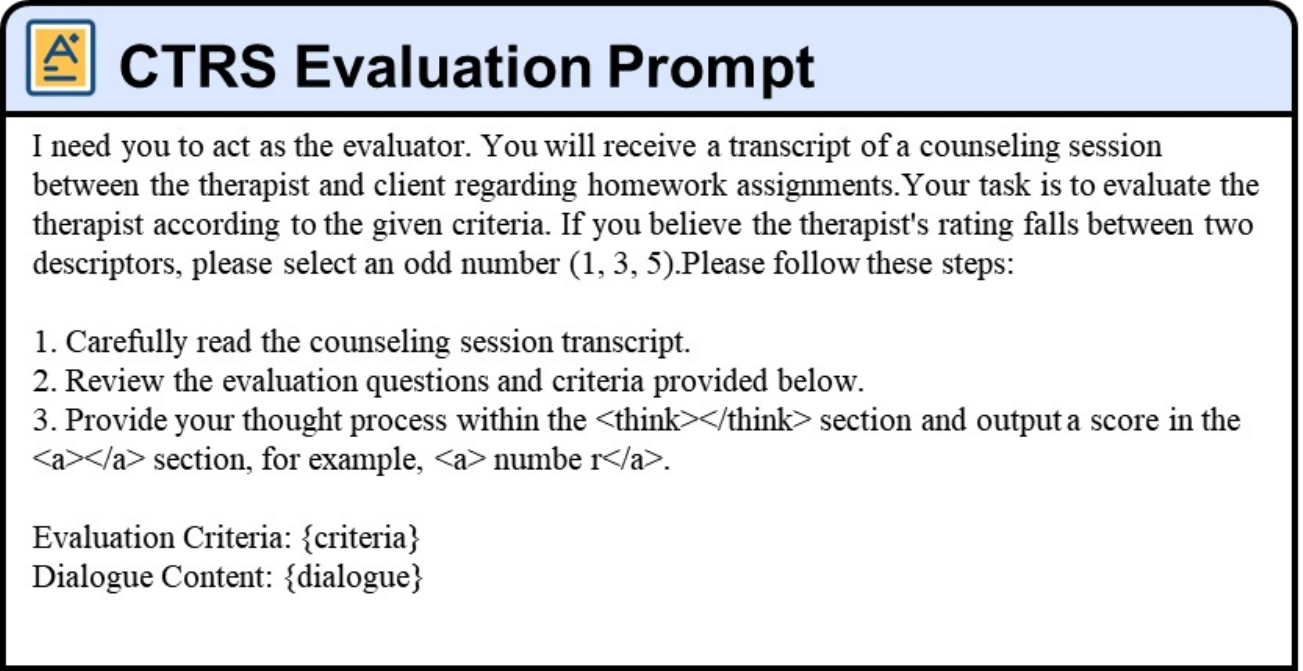}
  \caption{CTRS Evaluation Prompt.}
  \label{fig:CTRS}
\end{figure}

The Cognitive Therapy Rating Scale (CTRS) assesses both general counseling and CBT-specific competencies. The original scale includes six items for general skills (e.g., agenda setting, feedback, understanding, interpersonal effectiveness, collaboration, pacing, and time management) and six for CBT-specific skills (e.g., guided discovery, focus on key cognitions or behaviors, change strategies, technique application, and homework). For our evaluation, we selected a streamlined subset of six criteria that are reliably observable in text-based dialogues: \textbf{General skills}: Understanding, Interpersonal Effectiveness, Collaboration;\textbf{CBT-specific skills}: Guided Discovery, Focus, Strategy. The prompt used to guide CTRS scoring by gpt-4o-mini is shown in Figure~\ref{fig:CTRS}.

\begin{figure}[t]
  \includegraphics[width=\columnwidth]{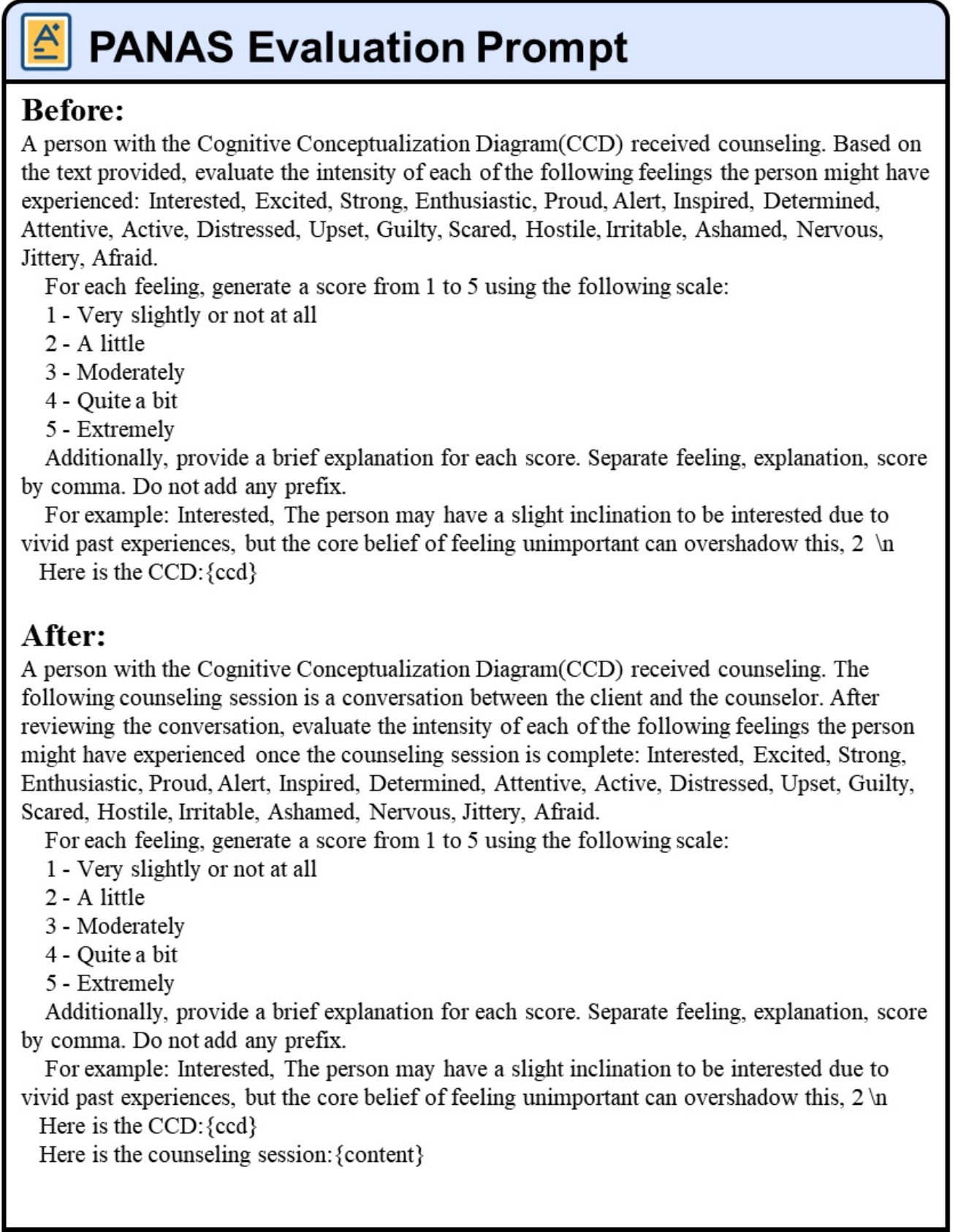}
  \caption{CTRS Evaluation Prompt.}
  \label{fig:panas}
\end{figure}

\begin{table*}
  \centering
  \small
  \begin{tabular}{lS[table-format=1.2]S[table-format=1.2]S[table-format=1.2]S[table-format=1.2]S[table-format=1.2]S[table-format=1.2]S[table-format=2.2]}
    \toprule
    \multirow{2}{*}{\textbf{Category}} & \multicolumn{3}{c}{\textbf{General Counseling Skills}} & \multicolumn{3}{c}{\textbf{CBT-specific Skills}} \\
    \cmidrule(r){2-4} \cmidrule(l){5-7}
    & \textbf{Understanding} & \textbf{Interpersonal Eff.} & \textbf{Collaboration} & \textbf{Guided Discovery} & \textbf{Focus} & \textbf{Strategy} & \\
    \midrule
    AVG. NEG      & 4.92 & 4.75 & 5.07 & 4.85 & 4.73 & 4.39 \\
    AVG. NEU   & 5.07 & 4.76 & \textbf{5.17} & \textbf{4.97} & \textbf{4.90} & \textbf{4.41} \\
    AVG. POS     & \textbf{5.11} & \textbf{4.82} & 5.16 & 4.85 & 4.88 & 4.36 \\
    \bottomrule
  \end{tabular}
  \caption{\label{tab:attitude}
    Impact of client attitude on simulator performance. 
  }
\end{table*}

\begin{table*}
  \centering
  \small
  \begin{tabular}{lS[table-format=1.2]S[table-format=1.2]S[table-format=1.2]S[table-format=1.2]S[table-format=1.2]S[table-format=1.2]}
    \toprule
    \multirow{2}{*}{\textbf{Category}} & \multicolumn{3}{c}{\textbf{General Counseling Skills}} & \multicolumn{3}{c}{\textbf{CBT-specific Skills}} \\
    \cmidrule(r){2-4} \cmidrule(l){5-7}
    & \textbf{Understanding} & \textbf{Interpersonal Eff.} & \textbf{Collaboration} & \textbf{Guided Discovery} & \textbf{Focus} & \textbf{Strategy} \\
    \midrule
    Helpless     & 5.03 & 4.73 & 5.13 & \textbf{4.93} & 4.83 & \textbf{4.43} \\
    Unlovable    & \textbf{5.04} & 4.75 & 5.11 & 4.81 & 4.83 & 4.35 \\
    Worthless    & 5.03 & \textbf{4.86} & \textbf{5.15} & 4.92 & \textbf{4.85} & 4.37 \\
    \bottomrule
  \end{tabular}
  \caption{\label{tab:core_belief}
    Impact of client's core belief on simulator performance.
  }
\end{table*}

\begin{table}
  \centering
  \small
  \begin{tabular}{lS[table-format=3.0]S[table-format=4.2]S[table-format=4.2]}
    \toprule
    \textbf{Category} & \textbf{Total} & \textbf{Client} & \textbf{Therapist} \\
     \midrule
    Cases & 4500 & {-} & {-} \\
    Characters per case & 4570.91 & 2901.64 & 1669.27 \\
    Characters per utterance & 281.27 & 178.55 & 102.72 \\
    \bottomrule
  \end{tabular}
  \caption{\label{tab:stats}
    Statistics of the \textsc{CCDChat} dataset.
  }
\end{table}

\begin{table*}[t]
  \centering
  \small
  \begin{tabular}{lS[table-format=1.2]S[table-format=1.2]S[table-format=1.2]S[table-format=1.2]S[table-format=1.2]S[table-format=1.2]}
    \toprule
    \multirow{2}{*}{\textbf{Datasets}} & \multicolumn{3}{c}{\textbf{General Counseling Skills}} & \multicolumn{3}{c}{\textbf{CBT-specific Skills}} \\
    \cmidrule(r){2-4} \cmidrule(l){5-7}
    & {\textbf{Understanding}} & {\textbf{Interpersonal Eff.}} & {\textbf{Collaboration}} & {\textbf{Guided Discovery}} & {\textbf{Focus}} & {\textbf{Strategy}} \\
    \midrule
    \textsc{CACTUS}       & 4.84 & 4.64 & 5.06 & 4.68 & 4.44 & 4.32 \\
    PsyDTCorpus  & 5.06 & 4.80 & 5.30 & 4.92 & {\textbf{4.92}} & 4.24 \\
    \textbf{\textsc{CCDChat}} & {\textbf{5.20}} & {\textbf{4.94}} & {\textbf{5.32}} & {\textbf{4.96}} & {\textbf{4.92}} & {\textbf{4.50}} \\
    \bottomrule
  \end{tabular}
  \caption{\label{tab:dataset_comparison}
    Automated CTRS evaluation of 150 randomly sampled cases across different datasets using GPT-4o-mini.
  }
\end{table*}

\subsection{Positive and Negative Affect Schedule (PANAS)}\label{app:panas}

The Positive and Negative Affect Schedule (PANAS) is a widely used self-report measure designed to assess the intensity of positive and negative affective states. It consists of two 10-item subscales: one for positive emotions (e.g., Excited, Strong, Enthusiastic, Proud, Alert, Inspired, Determined, Attentive, Active, Interest) and one for negative emotions (e.g., Distressed, Upset, Guilty, Scared, Hostile, Irritable, Ashamed, Nervous, Jittery, Afraid). Respondents rate each item on a 5-point scale from 1 (very slightly or not at all) to 5 (extremely), indicating how much they have experienced each emotion. In this study, PANAS is adapted as a pre- and post-counseling emotional assessment tool. The evaluation is conducted by LLM raters who infer the client's affective state before and after therapy based on the CCD and dialogue content. For each emotion, raters assign a score and provide a brief justification grounded in the clinical context. The structured prompt guiding this scoring process is shown in Figure~\ref{fig:panas}.

\begin{figure}[t]
  \includegraphics[width=\columnwidth]{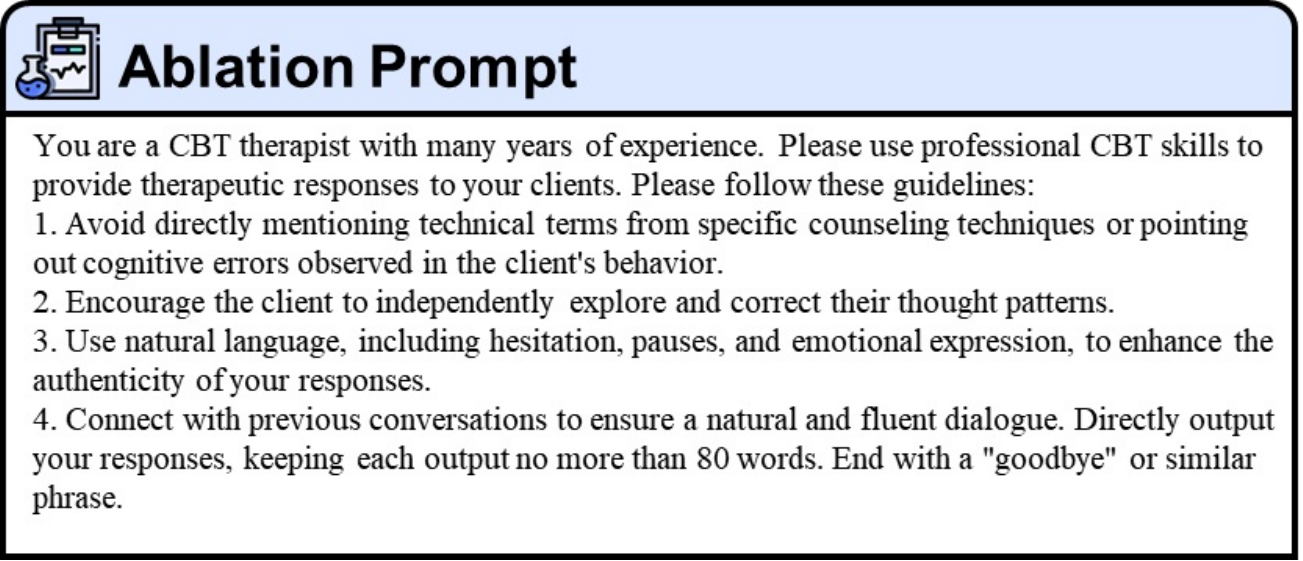}
  \caption{Ablation Experiment Prompt.}
  \label{fig:ablation}
\end{figure}

\section{Experimental Setup}

\subsection{Training}\label{app:training}
To minimize human-induced variability, we adopted trained models provided by ~\citet{lee2024cactus} and ~\citet{xie2025psydt}, selecting Qwen2.5-7B-Instruct as our base model due to its comparable parameter scale. We utilized the LoRA technique for fine-tuning with the assistance of the accelerate library \footnote{\url{https://hugging-face.cn/docs/accelerate}}on two NVIDIA A800-80GB GPUs in parallel. Identical hyperparameters were applied across all agents and ablation models. The training configuration is detailed below:

\begin{lstlisting}[language=Python, basicstyle=\ttfamily\small, frame=single, rulecolor=\color{black}, backgroundcolor=\color{gray!10}]
learning_rate=2e-5,
per_device_train_batch_size=2,
gradient_accumulation_steps=8,
gradient_checkpointing=True,  
num_train_epochs=3,
weight_decay=0.01,
lr_scheduler_type="cosine",
seed=42,
fp16=True
\end{lstlisting}

The prompt template used in the ablation study is illustrated in Figure~\ref{fig:ablation}.

\subsection{Inference} 
For all inference runs, we limited the maximum output length to 1024 tokens and applied a temperature of 0.7 to balance creativity and coherence. All remaining generation parameters were left at their default settings.

\section{Analysisi of \textsc{CCDChat}}
\subsection{Statistics} 

Constrained by API costs, we used a subset of the CCDs to generate \textsc{CCDChat} . While it demonstrates diversity and clinical grounding, it may not yet fully encapsulate the entire breadth of cognitive schemas encountered in practice. In future work, we plan to expand the dataset by incorporating additional CCD instances.

We report key average statistics of the \textsc{CCDChat} dataset in Table~\ref{tab:stats}. The dataset comprises 4500 dialogue cases.

\subsection{Impact of Client Attitude on Simulator Performance.} \label{app:attitude_analysis}
We randomly sampled 150 dialogues from each client attitude category (negative, neutral, and positive) and evaluated the simulator’s performance using GPT-4o-mini. The results are presented in Table~\ref{tab:attitude}.

\subsection{Impact of Client Core Belief on Simulator Performance.} \label{app:core_analysis}
We randomly sampled 150 dialogues for each core belief type (Helpless, Unlovable, Worthless) and evaluated simulator performance using GPT-4o-mini. As shown in Table~\ref{tab:core_belief}, scores across both general and CBT-specific skills remain consistently high. However, these differences are small in magnitude and do not reach statistical significance ($p > 0.05$). This suggests that the CCD-CBT framework is robust to variations in core belief content.

\subsection{Dataset Comparison and Clinical Completeness}

To rigorously evaluate the therapeutic capability of our proposed dataset against existing CBT datasets, we conducted both structural analyses and automated clinical evaluations.\\\\
\textbf{Structural Integrity and Phase Completeness.} A defining advantage of \textsc{CCDChat} is its strict adherence to clinical completeness. Professional CBT requires a highly structured, four-phase progression: Identification, Assessment, Intervention, and Summary. Our structural inspection reveals that baselines such as CACTUS and PsyDTCorpus frequently omit the critical "Assessment" phase, systematically failing to inquire about the client's degree of belief in their automatic thoughts. Furthermore, a quantitative analysis of PsyDTCorpus utilizing DeepSeek-V3 exposed a severe deficiency in the "Homework Assignment" module which is completely missing in 67.5\% (3,212 out of 4,760) of its cases. These structural omissions substantially degrade the professional fidelity of the simulated therapy. Conversely, \textsc{CCDChat} guarantees a structured, end-to-end therapeutic process complete with explicit homework planning.\\\\
\textbf{Quantitative Evaluation via CTRS.} We assessed the dialogue quality using the CTRS. We randomly sampled 150 dialogue instances from each dataset and utilized GPT-4o-mini to automatically score the therapeutic interactions. As detailed in Table~\ref{tab:dataset_comparison}, \textsc{CCDChat} consistently outperforms the baseline datasets across all core therapeutic dimensions.

\section{Cross-Model Validation and Evaluation Robustness}
Our experimental pipeline aligns with the established methodological flow adopted by recent high-quality frameworks, such as PsyDT and~\textsc{CACTUS}, which utilize GPT models for both data synthesis and dialogue evaluation. Recent research by Lee et al. empirically demonstrates that LLM-based evaluation methods possess the potential to effectively assess counseling efficacy, serving as a viable and scalable alternative to traditional expert evaluation~\citep{lee2024cactus}. 

However, relying solely on OpenAI models for evaluating OpenAI-generated text may introduce potential self-preference biases. To systematically mitigate this concern and rigorously validate the robustness of our simulated evaluation, we conducted an independent cross-model validation. We randomly sampled 150 interaction instances generated by three different models and employed Gemini-3-Flash—a completely distinct model architecture—as the independent evaluator.

This additional evaluation specifically focused on core CBT-specific skills. As shown in Table~\ref{tab:cross_model}, our proposed model maintains a significant performance advantage across all CBT metrics even when evaluated by a non-OpenAI judge. These cross-model results collectively and objectively demonstrate the intrinsic effectiveness of our proposed information-asymmetric architecture, proving that the therapeutic gains are architecturally driven rather than an artifact of evaluator bias.

\begin{table}
  \centering
  \small
  \begin{tabular}{lS[table-format=1.2]S[table-format=1.2]S[table-format=1.2]}
    \toprule
    \multirow{2}{*}{\textbf{Model}} & \multicolumn{3}{c}{\textbf{CBT-specific Skills}} \\
    \cmidrule(l){2-4}
    & {\textbf{Guided Discovery}} & {\textbf{Focus}} & {\textbf{Strategy}} \\
    \midrule
    Camel       & 4.78 & 5.03 & 4.90 \\
    SoulChat2.0 & 4.38 & 4.82 & 4.48 \\
    \textbf{Ours} & {\textbf{4.85}} & {\textbf{5.26}} & {\textbf{5.34}} \\
    \bottomrule
  \end{tabular}
  \caption{\label{tab:cross_model}
    Cross-model evaluation results on 150 randomly sampled instances using Gemini-3-Flash as the judge.
  }
\end{table}

\section{Case Study}
Figure~\ref{fig:ccd_diagram} presents the complete CCD of a client. Figures~\ref{fig:case_study_identification}–\ref{fig:case_study_summary} illustrate the full therapeutic process grounded in the CCD-CBT framework. Notably, the dialogues shown here are translated from the original Chinese. We observe that the CCD-CBT implementation tends to be slightly more verbose in Chinese than in English, likely due to linguistic and pragmatic differences between the two languages, such as the higher-context nature of Chinese communication and its preference for mitigated or elaborative phrasing in interpersonal discourse.

\begin{figure*}[htbp]
  \centering
  \fbox{
    \begin{minipage}{0.98\linewidth}
      \small
      \setlength{\parindent}{0pt}
      \setlength{\parskip}{4pt}

      \textbf{Situation:} Heavy workload coinciding with a cold. \\
      \textbf{Automatic Thought:} Why did I get sick now? My work won’t get done, I feel awful. I just want to cry. \\
      \textbf{Emotions:} Anxious, Disappointed, Angry \\
      \textbf{Behavior:} Pushing through illness, ignoring physical needs, attempting to work while unwell. \\
      \textbf{Coping Strategy:} Forcing oneself to work when sick to avoid feeling incapable. \\
      \textbf{Intermediate Belief:} I must always be highly efficient to prove my worth; being sick makes me seem incompetent. \\
      \textbf{Core Belief:} Worthless. \\
      \textbf{Relevant History:} Past experiences of intense deadline-related stress have amplified anxiety during physical illness.

    \end{minipage}
  }
  \caption{\label{fig:ccd_diagram}
    Example of Cognitive Conceptualization Diagram (CCD) for the client.
  }
\end{figure*}

\begin{figure*}[htbp]
  \centering
  \fbox{
    \begin{minipage}{0.98\linewidth}
      \small
      \setlength{\parindent}{0pt}
      \setlength{\parskip}{4pt}

      \textbf{Strategy 1:} First, inquire about the client's feelings and obtain the latest information.\\
      \textbf{Counselor:} Hello. If you'd like to share your recent feelings? Is there anything that's bothering you? \\
      \textbf{Client:} Hmm... I've been feeling quite tired lately. I have a lot of work, and then... I caught a cold. I feel like I chose the wrong time to get sick. I don't have much time left to complete my tasks, and now I'm not feeling well... I really don't know what to do. \\

      \textbf{Strategy 2:} The counselor identifies the specific context in which these emotional and behavioral problems exist.\\
      \textbf{Counselor:} It sounds like you're feeling a bit overwhelmed lately. Work and health pressures are piling up. Hmm… could you elaborate on what's causing your particular anxiety or disappointment? Is it time pressure, or worry about task completion? \\
      \textbf{Client:} Yes, I am a bit overwhelmed… I've been really busy at work lately, and I also caught a cold, which makes me feel even worse. I think it's probably because there are so many things to finish before deadlines, and I'm always worried that I won't be able to finish everything on time. And now that I'm sick, I feel even more hopeless and helpless. I know I should rest, but I also feel that resting might make me miss something, make me look less capable, or even unable to perform my job. \\

      \textbf{Strategy 4:} Starting with automatic thoughts, the therapist uses the downward arrow technique (e.g., ``If automatic thoughts actually occurred, what would that mean for you?'') to confirm the content of the client's core beliefs.\\
      \textbf{Counselor:} I hear your anxiety, especially your worry that taking a break will make you seem incompetent. This makes me think, what would it mean for you if this work wasn't completed on time? What might the impact be? Or what deeper anxieties might it trigger? \\
      \textbf{Client:} Sigh, I know I'm not feeling well, but I just feel like I can't stop. I'm so afraid people will think I'm letting them down because of my illness, or that I'm unreliable. If I take a break and the work isn't finished, what will people think of me? They might think I'm not capable of doing this job. I've always felt that if I don't perform well, my position and career development might be affected, and I'll be seen as useless. I just... don't want to be seen as an unreliable person. \\

      \textbf{Strategy 5:} By understanding the visitor's early parenting experiences, relevant information that shapes their core beliefs can be obtained.\\
      \textbf{Counselor:} It sounds like you have a deep concern about how others perceive you, especially putting a lot of pressure on yourself when you're not feeling well. I'm wondering if this emphasis on ability and reliability is related to your parents' expectations or evaluations of you in your early years? How do you think they influenced your self-perception? \\
      \textbf{Client:} Hmm… I think it might be related. I remember when I was little, my parents had very high expectations of me, especially regarding my studies and performance. They always wanted me to get good grades, and sometimes if I didn't meet their expectations, I felt like a failure. I think this might have made me feel that I had to perform exceptionally well to satisfy them and prove my worth. I tried very hard to meet their expectations, so now I always push myself to achieve goals at work, even when I'm sick. I think… that might be where it all comes from. \\

      \textbf{Strategy 7:} By examining relevant experiences in their personal growth history, we can identify their coping strategies (coping strategies refer to the strategies that clients use to conceal their core beliefs).\\
      \textbf{Counselor:} It sounds like you've internalized your parents' expectations as your own. Perhaps because of past experiences, you feel that you can only prove your worth through constant effort. Has the recent work and health pressure made these thoughts even more pronounced? Do you think there's a way to be less hard on yourself when you need rest? \\
      \textbf{Client:} Sigh, I'm not sure. I don't know why, but I always feel that if I'm not working efficiently, I'll be considered irresponsible. This feeling is especially strong when work pressure is high. I know I need rest when I'm sick, but every time I think about the consequences of stopping, I worry a lot. It's like if I don't push myself so hard, I... I'll seem like I'm not doing enough. I've thought about resting, but I just can't stop. Do you think there's any way to stop being bothered by these thoughts? \\

      \textbf{Strategy 8:} End of conceptualization phase.
    \end{minipage}
  }
  \caption{\label{fig:case_study_identification}
     Example of identification phase.
  }
\end{figure*}

\begin{figure*}[htbp]
  \centering
  \fbox{
    \begin{minipage}{0.98\linewidth}
      \small
      \setlength{\parindent}{0pt}
      \setlength{\parskip}{4pt}
    \textbf{Strategy 1:} Ask client about the intensity of their primary emotions, ranging from 0 to 100\%, with higher percentages indicating greater intensity.\\
    \textbf{Counselor:} I understand that this kind of worry can be very stressful. You mentioned that even when you're sick, you still feel the need to keep going. How intense are your anxieties during those moments? If you could give a percentage, what would it be? Perhaps we can explore some new ways of thinking to help you relax a little.  \\
    \textbf{Client:} Hmm... about 80\%, sometimes I even feel like I can't breathe. I know I'm sick, but there's always a voice inside telling me I can't stop, I have to keep going, otherwise I won't finish my work. The pressure can be so great that I feel completely exhausted.  \\
    
    \textbf{Strategy 2:} Assess the degree of belief of the client's automatic thoughts, ranging from 0 to 100\%, with higher percentages indicating greater belief.\\
    \textbf{Counselor:} It sounds like you're under a lot of pressure, especially when you're sick. You mentioned having to keep working, so how strong is that thought when it arises? We could try looking at this from a different angle; maybe that would alleviate some of your stress.\\
    \textbf{Client:} Hmm, about 90\%. Because if I don't keep working, things pile up, and then it might be harder for me to complete tasks, and I might be seen as irresponsible or incompetent. It's like I can only prove my worth by constantly and efficiently completing my work. So, whenever I'm sick, I'm always particularly worried about not finishing my work, and I even become more anxious. Although I understand that rest might be good for my health, I just can't completely relax and rest.\\

    \textbf{Strategy 3:} End of assessment phase.
    \end{minipage}
  }
  \caption{\label{fig:case_study_assessment}
     Example of assessment phase.
  }
\end{figure*}

\begin{figure*}[htbp]
  \centering
  \fbox{
    \begin{minipage}{0.98\linewidth}
      \small
      \setlength{\parindent}{0pt}
      \setlength{\parskip}{4pt}
    \textbf{Strategy 1:} Guide client to provide evidence supporting and opposing automatic thoughts.\\
    \textbf{Counselor:} Well, I understand your concerns; these thoughts can be hard to shake off. However, resting when you're sick is actually part of enhancing your work performance. Is it possible that, to some extent, you can only be more efficient at work after a good rest? Have you ever felt more energetic after resting to tackle a backlog of tasks? Such experiences might offer some new insights.  \\
    \textbf{Client:} Hmm...sometimes rest does make me feel better, and then I have more energy when I continue working. Like last time, after sleeping all weekend, my work efficiency on Monday improved significantly. However, when I'm really unwell, I always feel I can't waste time resting, especially when tasks are piling up. This contradiction makes me very irritable...I always feel that getting sick is bad news that disrupts my plans. I may know that rest has benefits, but it's just hard to completely let go and accept this fact.  \\
    
    \textbf{Strategy 2:} Offer the client alternative thoughts that are the opposite of their automatic thoughts and guide the client to provide evidence supporting or opposing the alternative thoughts.\\
    \textbf{Counselor:} I understand how you feel; balancing rest and work is very difficult when under pressure. But you mentioned increased efficiency after rest, which might suggest that rest isn't a waste of time but a strategy to improve ability. So, if rest is another way to maintain efficiency, what examples can you think of that support or oppose it? This might help you accept rest as part of work more readily.\\
    \textbf{Client:} Hmm... For example, to support this, I inexplicably insisted on working after catching a cold a few times, which only made my symptoms worse and reduced my efficiency. On the contrary, after occasionally resting and adjusting, I felt more focused and motivated to handle the backlog of tasks. As for the opposing example, it's probably my concern. Every time I stop to rest, I feel like I'm wasting time, especially when there are many tasks waiting to be done. Perhaps I still habitually believe that efficiency means constantly being busy. Although I understand that this isn't necessarily true, changing this mindset is really difficult... 

    \textbf{Strategy 3:} Ask the client how much they believe in the alternative thinking.\\
    \textbf{Counselor:} Well, so how much do you now believe that taking a break is another way to stay productive?\\
    \textbf{Client:} Yeah, maybe that's true, but I still find it hard to rest without thinking about anything. Is there any way to make it easier for me to accept rest when facing this contradiction?\\

    \textbf{Strategy 4:} Design detailed and executable behavioral experiments to validate the findings, and encourage client to try them.\\
    \textbf{Counselor:} Perhaps you could start with an experimental mindset, setting a short break, like 10 minutes, and observing how you feel working afterward. Do you think this fine-tuning would give you more confidence?\\
    \textbf{Client:} I think trying a 10-minute break is a good start. Such a short break won't make me feel like I'm wasting too much time. I can try it first and see if I can better recover my energy in such a short period.\\

    \textbf{Strategy 5:} End of intervention phase.
    
    \end{minipage}
 
  }
  \caption{\label{fig:case_study_intervention}
     Example of intervention phase.
  }
\end{figure*}

\begin{figure*}[htbp]
  \centering
  \fbox{
    \begin{minipage}{0.98\linewidth}
      \small
      \setlength{\parindent}{0pt}
      \setlength{\parskip}{4pt}
    \textbf{Strategy 1:} Summarize the discussed homework with the client: \{Try setting a 10-minute break and see if you can better restore your energy in that short time, and observe whether it helps improve work efficiency.\} Avoid using terms like "homework" or "task" directly; use terms like "practice," "experiment," or "proof experiment."\\
    \textbf{Counselor:} Great, it sounds like you're ready to try this little experiment. Try using this time as an opportunity for self-observation and see if you feel any changes after the short break. Perhaps you'll discover some unexpected positive effects in practice. Remember to observe changes in your state and don't overlook those small improvements. Keep it up!\\
    \textbf{Client:} Okay, I'll try my best. Although I'm still a little uneasy, worried that it might affect my work progress, I want to try and see what changes this short break can bring. Thank you for your suggestion. I will try to remember my small improvements and find more motivation from them.\\

    \textbf{Strategy 2:} Provide closing statements that express affirmation and encouragement to the visitor. End with a "goodbye" or similar phrase.\\
    \textbf{Counselor:} Your courage is commendable. Acknowledging your unease and continuing to try is the starting point for change. Remember, every small step forward is an important step towards a healthier work rhythm. Keep an open mind, observe the changes after your rest, and hopefully you will see the motivation that progress brings. Good luck, and I look forward to hearing your experience sharing. Goodbye!\\
    \textbf{Client:} Thank you, I will continue to try. I know this won't happen overnight, but I will try to keep an open mind to these new attempts and see if I can find some positive changes from them. Thank you for your encouragement, see you next time!\\

    \textbf{Strategy 3:} End of summary phase.

      \end{minipage}
 
  }
  \caption{\label{fig:case_study_summary}
     Example of summary phase.
  }
\end{figure*}

\end{document}